\pdfoutput=1

\documentclass[11pt]{article}

\usepackage[final]{acl}
\usepackage{xcolor}

\usepackage{times}
\usepackage{latexsym}
\usepackage{multirow}
\usepackage{diagbox}
\usepackage[T1]{fontenc}

\usepackage[utf8]{inputenc}

\usepackage{microtype}

\usepackage{inconsolata}

\usepackage{graphicx}
\usepackage{newfloat}
\usepackage{listings}
\usepackage{colortbl}
\usepackage{algorithm}
\usepackage{algorithmic}
\usepackage{multirow}
\usepackage{booktabs}
\usepackage{amsmath}
\usepackage{subfigure}
\usepackage{longtable}

\DeclareCaptionStyle{ruled}{labelfont=normalfont,labelsep=colon,strut=off} 
\floatstyle{ruled}
\newfloat{listing}{tb}{lst}{}
\floatname{listing}{Listing}
\newcommand{\eg}{\hbox{\emph{e.g.}}}
\newcommand{\ie}{\hbox{\emph{i.e.}}}
\newcommand{\etc}{\hbox{\emph{etc.}}~}

\newcommand{\name}{\textsc{CruxEval-X}}

\newcommand{\inputtask}{input reasoning}
\newcommand{\outputtask}{output reasoning}

\title{\name: A Benchmark for Multilingual Code \\ Reasoning, Understanding and Execution}

\author{
    Ruiyang Xu\textsuperscript{\rm 1,\rm 2,}\textsuperscript{*},
    Jialun Cao\textsuperscript{\rm 3,}\textsuperscript{*},
    Yaojie Lu\textsuperscript{\rm 1},
    Ming Wen\textsuperscript{\rm 4},
    Hongyu Lin\textsuperscript{\rm 1},\\
    \textbf{Xianpei Han}\textsuperscript{\rm 1},
    \textbf{Ben He}\textsuperscript{\rm 1,\rm 2},
    \textbf{Shing-Chi Cheung}\textsuperscript{\rm 3},
    \textbf{Le Sun}\textsuperscript{\rm 1}
\\
\normalsize
\textsuperscript{\rm 1}Chinese Information Processing Laboratory, Institute of Software, Chinese Academy of Sciences \\
\normalsize
    \textsuperscript{\rm 2}University of Chinese Academy of Sciences 
\normalsize
    \textsuperscript{\rm 3}The Hong Kong University of Science and Technology \\ 
\normalsize
    \textsuperscript{\rm 4}Huazhong University of Science and Technology\\
\normalsize
    \texttt{\{xuruiyang2022,luyaojie,hongyu,xianpei,sunle\}@iscas.ac.cn} \\
\normalsize
    \texttt{\{jcaoap, scc\}@cse.ust.hk mwenaa@hust.edu.cn benhe@ucas.edu.cn}
}

\definecolor{codegreen}{rgb}{0,0.6,0}
\definecolor{codegray}{rgb}{0.5,0.5,0.5}
\definecolor{codepurple}{rgb}{0.58,0,0.82}
\definecolor{backcolour}{rgb}{0.95,0.95,0.92}
\definecolor{darkgreen}{rgb}{0,0.5,0}

\lstdefinestyle{mystyle}{   
    commentstyle=\color{codegreen},
    keywordstyle=\color{magenta},
    numberstyle=\color{codegray},
    stringstyle=\color{codepurple},
    morekeywords={define, sort, take},
    morecomment=[l]{;;}
}

\begin{document}
\maketitle
\renewcommand{\thefootnote}{}
\footnotetext{\textsuperscript{*}These authors contributed equally to this work.}
\renewcommand{\thefootnote}{\arabic{footnote}}

\begin{abstract}
Code benchmarks such as HumanEval are widely adopted to evaluate Large Language Models' (LLMs) coding capabilities. However, there is an unignorable \textbf{\textit{programming language bias}} in existing code benchmarks -- over 95\% code generation benchmarks are dominated by Python, leaving the LLMs' capabilities in other programming languages such as Java and C/C++ unknown. Moreover, \textbf{\textit{coding task bias}} is also crucial. Most benchmarks focus on code generation capability, while benchmarks for \textit{code reasoning} (given input, reasoning output; and given output, reasoning input), an essential coding capability, are insufficient. 
Yet, constructing multi-lingual benchmarks can be expensive and labor-intensive, and codes in contest websites such as Leetcode suffer from data contamination during training. To fill this gap, we propose \name, a \textbf{\textit{multi-lingual code reasoning benchmark}} that contains 19 programming languages. It comprises at least 600 subjects for each language, along with 19K content-consistent tests in total. In particular, the construction pipeline of \name~works in a fully automated and test-guided manner, which iteratively generates and repairs based on execution feedback. Also, to cross language barriers (e.g., dynamic/static type systems in Python/C++), we formulated various transition rules between language pairs to facilitate translation. 
Our extensive evaluation of 24 representative LLMs reveals the correlation between language pairs. For example, TypeScript and JavaScript show a significant positive correlation, while Racket has less correlation with other languages. More interestingly, even a model trained solely on Python can achieve at most 34.4\% Pass@1 in other languages, revealing the cross-language generalization of LLMs.
\end{abstract}

\section{Introduction}
\begin{figure}[ht!]
\centering
\includegraphics[width=\linewidth]{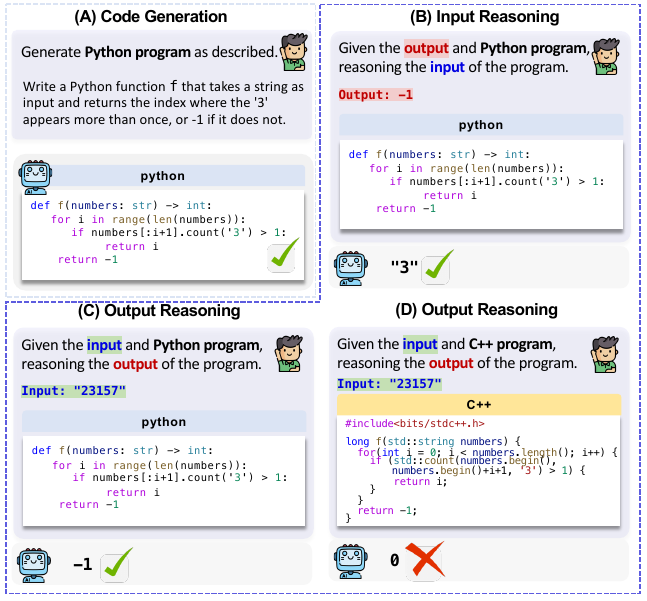}
\caption{Code generation vs. Code reasoning}
\label{fig:example}
\end{figure}

{Large language models (LLMs) have shown advanced proficiency in various domains, including code generation~\citep{liu2024your,du2024evaluating}, defect detection~\citep{yang2024large} and program repair~\citep{xia2023keep,zhong2024ldb,hu2024leveraging}. Benchmarks such as HumanEval~\citep{chen2021evaluating} and SWE-bench~\citep{jimenez2023swe} were introduced to measure LLMs' capabilities, providing insights into LLMs' strengths and weaknesses. }

{Recent studies~\citep{javabench,chai2024mceval,chen2024reasoningruntimebehaviorprogram} have spotted two significant biases in current benchmarks. First, \textbf{\textit{Programming language bias}}. As pointed out by prior studies~\citep{javabench,chai2024mceval,chen2021evaluating,austin2021program}, \textit{Python} dominates code generation benchmarks with over 95\% involvement. Other programming languages (PLs) such as Java and C/C++, despite their popularity and availability, gain less exploration. Second, \textbf{\textit{Coding task bias}}. Most coding benchmarks focus on code generation tasks (\ie, giving descriptions in natural language and generating the program, as shown in Figure~\ref{fig:example} (A)), while \textbf{\textit{code reasoning}} (\ie, given the program, reasoning the input or output of the program, as shown in Figure~\ref{fig:example} (B - D)) is seldom evaluated~\citep{chen2024reasoningruntimebehaviorprogram}.
}
A recent work introduced a code reasoning benchmark~\citep{gu2024cruxeval,chen2024reasoningruntimebehaviorprogram}, while it is only in Python. Figure~\ref{fig:example} (C - D) shows that simply changing PLs from Python to C++ can turn a correct reasoning into an incorrect one.  {More discussion about this task can be found in Appendx \ref{sec:J}}. 

{However, constructing multi-lingual benchmarks is not a trivial task. First, \textbf{\textit{human annotation can be expensive}}. As reported by recent work~\citep{chai2024mceval}, they spent a total of \$12,000 US dollars for human annotators, providing the working environment, free meals, souvenirs, and free GPT-4 interface usage to construct their multi-lingual benchmark. Second, \textbf{\textit{automated translation does not perform well}}. The latest studies~\citep{yin2024rectifiercodetranslationcorrector} show that even the best LLM (\ie, ChatGPT) can only achieve an average of 64\% success translation rate, which is far from practice. Rule-based translation~\citep{cassano2023multipl,inRustWeTrust22} usually suffers from generalizability issues, making them limited in handling prescribed code structures. Additionally, multi-lingual solutions from contest websites such as LeetCode and Codeforces were included in most LLMs training sources, thus suffering from \textbf{\textit{data contamination issues}}~\citep{cao2024concerneddatacontamination}. 
}

To fill the above research gaps, we introduce \name, a multi-lingual code reasoning benchmark that contains 19 popular PLs, including C++, Rust, Java, \etc, expanded from CruxEval~\citep{gu2024cruxeval}, a code reasoning benchmark written in Python. For each PL in \name, there are at least 600+ functions. In total, there are 12,660 subjects along with 19K test cases for input/output reasoning. 

Noteworthy that the pipeline of constructing \name~works in a fully automated manner. It first translates the test cases by transition rules adapted from prior work~\citep{cassano2023multipl}, then iterates the generation-and-repair process intensively. In particular, the transition rules are formulated to cross the language barriers. For example, Python employs a dynamically typed system where types are determined at runtime, whereas C++ uses a statically typed system requiring explicit type declarations at compile time. The rules facilitate the translation of the test cases. Additionally, inspired by prior work~\citep{yin2024rectifiercodetranslationcorrector,testGuidedTranslation}, we employ a test-guided manner~\citep{testGuidedTranslation} to generate the translation and iteratively repair the generated code using execution feedback (\eg, compilation error, runtime error)~\citep{yin2024rectifiercodetranslationcorrector}. 

Through extensive experiments on 24 mainstream LLMs, we observe several interesting findings. First, in multiple PLs, the input reasoning and output reasoning capabilities of LLMs are comparable. Also, there is a noticeable correlation between certain PLs (\eg, JavaScript and TypeScript show a positive correlation, while Racket consistently yields the worst results). More interestingly, we observe that even if a model is only trained on Python (e.g., phi-1 and phi-1.5), it still can reach a 16\% $\sim$ 26\% output reasoning success rate in other PLs, compared with 25.6\% in Python. The finding indicates the cross-language generalization of LLMs.

The contributions can be summarized as follows: (1) We introduce \name, a multi-lingual code reasoning benchmark that contains 19 popular PLs. (2) We introduce an automated code translation pipeline that adopts a test-guided and iterative generate-and-repair practice.
(3) We evaluate 20+ LLMs 
and yield inspiring findings.
\begin{figure*}[ht]
\centering
\includegraphics[width=\textwidth]{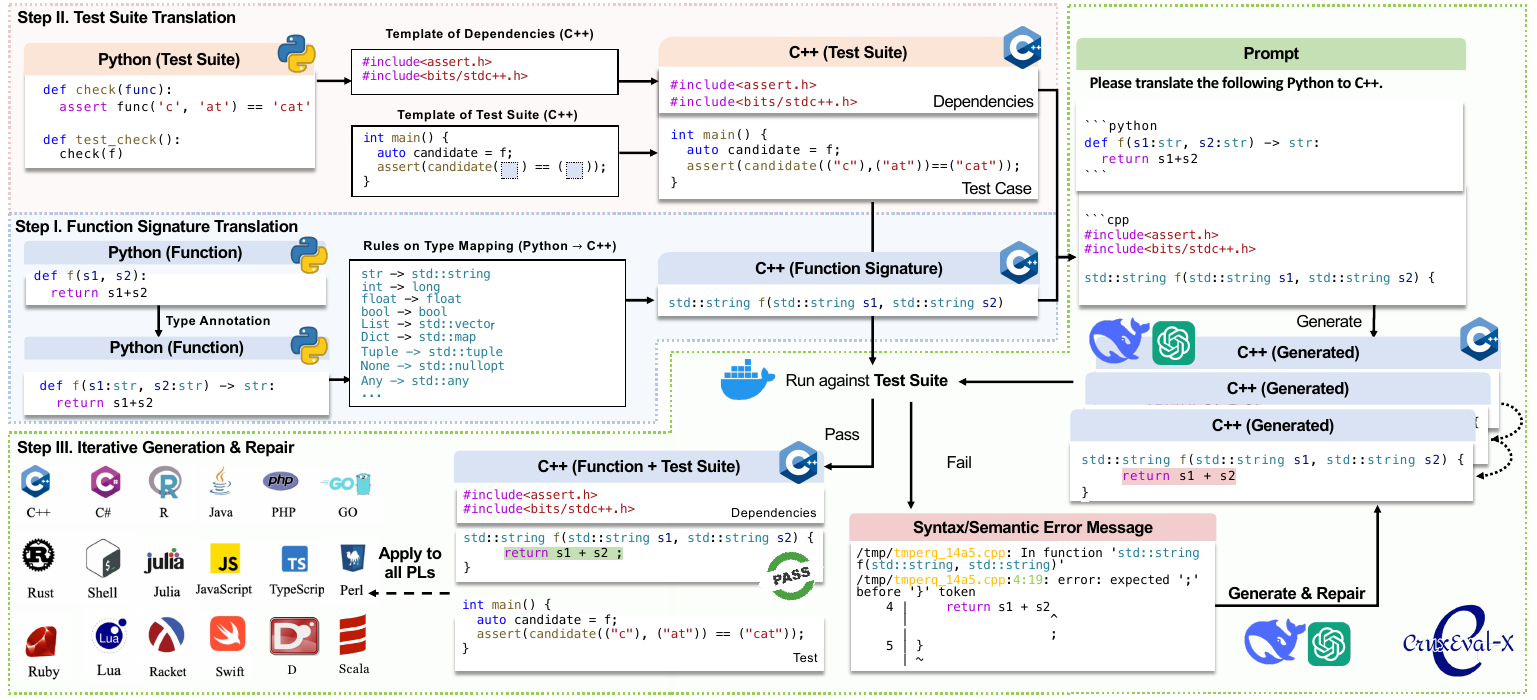}
\caption{The pipeline of CruxEval-X construction.}
\label{fig:workflow}
\end{figure*}
\section{Benchmark Construction}

In this section, we detailed the construction process of \name~in Figure \ref{fig:workflow}. It can be divided into three main steps. First, we translate function signatures via mapping variable type annotations (Step 1 in Figure \ref{fig:workflow}). Then we employ a rule-based approach to translate Python test cases into other PLs (Step 2 in Figure \ref{fig:workflow}). Finally, we integrate multiple LLMs to translate code by iterating the generation-and-repair process (Step 3 in Figure \ref{fig:workflow}).

\subsection{Step I. Function Signature Translation}
To enhance the accuracy and standardization of function translation results, we first translate the function signatures and dependencies.
Note that Python does not require an explicit type annotation, which may confuse the translation for the function signature. For example, as shown in Figure~\ref{fig:workflow} Step I, the types of two input parameters (\ie, \texttt{s1} and \texttt{s2}) are unclear. So we extract the input variables from the function signature using the syntax tree and match them with the tests. Based on the variable types in the tests, we annotate the input and output variables in the function signature.

Then, we adopt the rules as prior work~\citep{cassano2023multipl} to map the types from Python to other PLs. In particular, we identify the data types in the annotated Python signature (\eg, parameter types, return types), mapping the types from Python to other PLs according to the rules, then structuring the signature in the corresponding PLs. Take the example in Figure~\ref{fig:workflow} Step I, the Python signature \texttt{def f(s1:str, s2:str) -> str} is translated into that in C++ (\texttt{std::string f(std::string s1, std::string s2)}. After translating the tests, all 800 subjects in Python can be translated, as shown in Table~\ref{tab:pipeline}.

\subsection{Step II. Test Suites Translation}
We employ a test-guided approach to ensure the correctness of the translation results, which necessitates test cases in various PLs. Prior works~\citep{athiwaratkun2022multi,cassano2023multipl} provided various rules for mapping Python test cases to other PLs. We adopt the mapping rules from MutiPL-E~\citep{cassano2023multipl} to assist the transition of our test suites.

However, their rules have limited support for type handling (\eg, they cannot handle a list with hybrid types), 
Thus, to maximize the success rate, we made two improvements to enhance the rules.  
First, we enhance handling structured types such as \texttt{List} and \texttt{Dict}. For example, when handling C\#, we add an equality function to check whether two \texttt{Dict} types are equal.
Second, when dealing with variables that have complex types that are not as well-supported in some other PLs, we transform these variables into more generic types without significantly altering the original function's functionality. For example, we change type \texttt{List[Union(str, int)]} into \texttt{List(str)} if the function keeps the same functionality. A small portion of the data that cannot be converted is discarded. The result of Step II is shown in Table~\ref{tab:pipeline}. Further details can be found in the Appendix \ref{sec:B}.

\subsection{Step III. Iterative Generation \& Repair}
After translating tests, dependencies, and signatures, we employ multiple LLMs to iteratively translate Python code into target PLs. Each LLM performs two steps: generation and repair. Problems that pass all test cases are excluded from subsequent iterations. Below, we will elaborate on these two substeps in detail.

\subsubsection{Generation.}
Relying on a single LLM's limited generations often results in a low translation success rate~\citep{yin2024rectifiercodetranslationcorrector}. To address this, we propose a multi-round generation method that interacts with the testing environment to determine whether to continue iterating.

Let $A_0$ denote the number of correctly translated codes, and $U$ the total number of problems. For unresolved problems, LLM $M$ generates results over multiple rounds. Let $A_i$ represent the number of correct results in the $i$-th round. The maximum rounds is $N$. Early stopping occurs if the increase in correct results compared to $k$ rounds prior, $A_{i-k}$, falls below a threshold $\delta$. The formula is as follows:
\begin{equation}
\begin{aligned}
A_i = Correct&(P(O_{i}\ |\ U - A_{i-1}; M)) + A_{i-1} \\
\text{for }& i \in \{1, 2, \ldots, N\} \\
\text{Stop if } (i >& k) \text{ and } (A_i - A_{i-k} < \delta)
\end{aligned}
\end{equation}
Here, $O_i$ is the $i$-th round code generated from $P(O_{i}\ |\ U - A_{i-1}; M)$, and $Correct(.)$ calculate the correct results. To leverage diverse LLMs' strengths, we use GPT-3.5-Turbo for initial generation (see Table \ref{tab:pipeline}, ``w/o Iter'') and DeepseekCoder-33B-Instruct for further generation. This pipeline improves upon single-LLM generation, as demonstrated in our final results. (see Table \ref{tab:pipeline}, ``w/ Iter'') 
\subsubsection{Repair}
Simply generating code will still result in many errors that LLMs cannot solve. Therefore, after the generation step of each LLM, we provide them with error messages for error correction. Multiple iterative error correction is costly with limited benefits.~\citep{chen2023teaching}, so we only perform error correction once after each LLMs.

After GPT3.5-Turbo generation, we directly provide the LLM with the erroneous code along with the error messages for correction. After the Iterating Generation of DeepseekCoder-33B-Instruct, since the untranslated code can produce numerous incorrect code snippets after multiple rounds of iteration, which may contain the same errors. We first use simhash to deduplicate the erroneous code and then proceed with error correction on the deduplicated code. The error correction in each phase uses the LLM used in that phase.
with a temperature setting of 0.

\subsubsection{Multiturn Repair Based on Overlap}
After completing the steps above, we first calculate the intersection of correctly answered questions in different PLs. To our surprise, there were only 333 questions that all PLs answered correctly. However, 563 problems have been successfully translated correctly by at least 16 PLs. Upon analyzing the questions our LLM failed to solve, we find that each PL has its difficulties in translating from {Python}. For example, in Julia, the index for arrays and other collection types starts from 1, which differs from Python. The details of the difficulties can be found in Appendix \ref{sec:C}.

Based on these observations, we conduct final generation and iterative error correction on the questions that are correctly translated by more than 15 PLs. Due to the small amount of data, we utilized GPT-4o for generation and error correction. During the generation process, we provide the LLM with three corresponding typical examples based on the difficulties we find. The overlap is increased to 462 after repair of GPT-4o. 

We manually refined 38 questions that GPT-4 nearly solved, expanding our dataset to 500 entries. We determined that 500 entries are sufficient to distinguish the effectiveness of the LLMs. Therefore, we use these 500 entries as our \name~benchmark. The final result of our pipeline is shown in Table \ref{tab:pipeline} under the column ``w/ Iter''. The experimental setups and prompts can be found in Appendix \ref{sec:D},
{and detail statics of our benchmark is shown in Appendix \ref{sec:I}}.

\subsection{Quality Analysis}
After constructing the dataset, we evaluate the benchmark's quality from four perspectives. {First, \textbf{\textit{Accuracy}}: We use a test-guided approach to ensure that each data point passes test cases, guaranteeing correctness. Second, \textbf{\textit{Reliability}}: We build on the high-quality CRUXEVAL dataset, ensuring problems avoid randomness or multi-threading, enhancing reliability.} Third, \textbf{\textit{Data Leakage}}: We compare with Stack v2 (67.5TB of GitHub data), finding only 0.8\% overlap, indicating minimal leakage risk. Fourth, \textbf{\textit{Data Bias}}: We assess potential bias by generating evaluation data with other LLMs using the same method as described above. Results show no significant advantage for any LLM on its own generated data. The possibel reason is the distinct tasks in generation and evaluation phases. Detailed results and analysis are in Appendix \ref{sec:A}.


\begin{table}[ht]
  \small
    \begin{tabular}{l|c|c|c|c}
    \toprule
    \multirow{2}{*}{\textbf{Languages}} & \multirow{2}{*}{\textbf{Step I}} & \multirow{2}{*}{\textbf{Step II}} & \multicolumn{2}{c}{\textbf{Step III}} \\
     &       &       & \multicolumn{1}{c}{\textbf{w/o Iter}} & \textbf{w/ Iter} \\
    \midrule
    C\# (cs) & 800   & 774   & 380   & 670 \\
    C++ (cpp) & 800   & 800   & 549   & 733 \\
    D (d) & 800   & 754   & 95    & 629 \\
    GO (go) & 800   & 752   & 293   & 699 \\
    Java (java) & 800   & 774   & 541   & 698 \\
    JavaScript (js) & 800   & 800   & 634   & 743 \\
    Julia (jl) & 800   & 774   & 410   & 680 \\
    Lua (lua) & 800   & 800   & 582   & 741 \\
    Perl (pl) & 800   & 799   & 591   & 728 \\
    PHP (php) & 800   & 800   & 622   & 755 \\
    R (r) & 800   & 800   & 542   & 699 \\
    Racket (rkt) & 800   & 800   & 264   & 681 \\
    Ruby (rb) & 800   & 800   & 658   & 748 \\
    Rust (rs) & 800   & 754   & 449   & 690 \\
    Scala (scala) & 800   & 799   & 462   & 712 \\
    Shell (sh) & 800   & 763   & 528   & 674 \\
    Swift (swift) & 800   & 796   & 415   & 654 \\
    TypeScript (ts) & 800   & 774   & 592   & 726 \\
    \bottomrule
    \end{tabular}%
      \caption{The result of each step, The portion within parentheses in the ``Language'' column represents the abbreviations for various languages. Due to the constraints of page size, these abbreviations are used to better display certain charts or tables.}
  \label{tab:pipeline}%
\end{table}%

\section{Experiments}
\subsection{Experiment Setup}

\begin{table*}[h!]
  \small
  \setlength{\tabcolsep}{0.85mm}
  \resizebox{1\textwidth}{!}{
    \begin{tabular}{l|c|ccccccccccccccccccc}
    \toprule
    \multicolumn{21}{c}{\textbf{Input Reasoning Performence}} \\
    \midrule
    \textbf{Models} & \textbf{Size} & \multicolumn{1}{c}{\textbf{cs}} & \multicolumn{1}{c}{\textbf{cpp}} & \multicolumn{1}{c}{\textbf{d}} & \multicolumn{1}{c}{\textbf{go}} & \multicolumn{1}{c}{\textbf{java}} & \multicolumn{1}{c}{\textbf{js}} & \multicolumn{1}{c}{\textbf{jl}} & \multicolumn{1}{c}{\textbf{lua}} & \multicolumn{1}{c}{\textbf{pl}} & \multicolumn{1}{c}{\textbf{php}} & \multicolumn{1}{c}{\textbf{py}} & \multicolumn{1}{c}{\textbf{r}} & \multicolumn{1}{c}{\textbf{rkt}} & \multicolumn{1}{c}{\textbf{rb}} & \multicolumn{1}{c}{\textbf{rs}} & \multicolumn{1}{c}{\textbf{scala}} & \multicolumn{1}{c}{\textbf{sh}} & \multicolumn{1}{c}{\textbf{swift}} & \multicolumn{1}{c}{\textbf{ts}} \\
    \midrule
    GPT-4o & -     & \cellcolor[rgb]{ .557,  .686,  .851} 70.2 & \cellcolor[rgb]{ .592,  .71,  .863} 64.6 & \cellcolor[rgb]{ .549,  .678,  .847} 71.6 & \cellcolor[rgb]{ .525,  .663,  .839} 75.4 & \cellcolor[rgb]{ .561,  .686,  .851} 69.8 & \cellcolor[rgb]{ .541,  .675,  .843} 73.2 & \cellcolor[rgb]{ .58,  .702,  .855} 67.0 & \cellcolor[rgb]{ .541,  .675,  .843} 73.0 & \cellcolor[rgb]{ .557,  .686,  .851} 70.2 & \cellcolor[rgb]{ .529,  .667,  .839} 74.8 & \cellcolor[rgb]{ .557,  .686,  .847} 70.6 & \cellcolor[rgb]{ .533,  .667,  .839} 74.4 & \cellcolor[rgb]{ .576,  .698,  .855} 67.4 & \cellcolor[rgb]{ .549,  .678,  .847} 72.0 & \cellcolor[rgb]{ .537,  .671,  .843} 73.6 & \cellcolor[rgb]{ .588,  .71,  .859} 65.4 & \cellcolor[rgb]{ .557,  .686,  .847} 70.6 & \cellcolor[rgb]{ .533,  .667,  .839} 74.2 & \cellcolor[rgb]{ .533,  .671,  .839} 74.0 \\
    GPT-4o-mini & -     & \cellcolor[rgb]{ .631,  .737,  .875} 58.8 & \cellcolor[rgb]{ .675,  .769,  .89} 52.2 & \cellcolor[rgb]{ .62,  .729,  .871} 60.6 & \cellcolor[rgb]{ .612,  .722,  .867} 62.0 & \cellcolor[rgb]{ .639,  .745,  .878} 57.2 & \cellcolor[rgb]{ .627,  .733,  .875} 59.6 & \cellcolor[rgb]{ .647,  .749,  .878} 56.2 & \cellcolor[rgb]{ .6,  .718,  .863} 63.4 & \cellcolor[rgb]{ .639,  .745,  .878} 57.4 & \cellcolor[rgb]{ .616,  .729,  .871} 61.0 & \cellcolor[rgb]{ .627,  .733,  .875} 59.6 & \cellcolor[rgb]{ .62,  .729,  .871} 60.4 & \cellcolor[rgb]{ .678,  .773,  .89} 51.2 & \cellcolor[rgb]{ .612,  .725,  .867} 61.6 & \cellcolor[rgb]{ .616,  .725,  .871} 61.2 & \cellcolor[rgb]{ .671,  .765,  .886} 52.6 & \cellcolor[rgb]{ .639,  .745,  .878} 57.2 & \cellcolor[rgb]{ .6,  .718,  .863} 63.4 & \cellcolor[rgb]{ .616,  .725,  .871} 61.2 \\
    GPT-3.5 Turbo & -     & \cellcolor[rgb]{ .675,  .769,  .89} 52.2 & \cellcolor[rgb]{ .757,  .827,  .918} 39.2 & \cellcolor[rgb]{ .686,  .776,  .894} 50.2 & \cellcolor[rgb]{ .667,  .761,  .886} 53.4 & \cellcolor[rgb]{ .651,  .753,  .882} 55.4 & \cellcolor[rgb]{ .686,  .776,  .894} 50.0 & \cellcolor[rgb]{ .706,  .792,  .898} 47.0 & \cellcolor[rgb]{ .667,  .765,  .886} 53.2 & \cellcolor[rgb]{ .702,  .788,  .898} 47.6 & \cellcolor[rgb]{ .675,  .769,  .89} 52.2 & \cellcolor[rgb]{ .675,  .769,  .89} 51.6 & \cellcolor[rgb]{ .694,  .784,  .898} 48.6 & \cellcolor[rgb]{ .718,  .8,  .902} 45.4 & \cellcolor[rgb]{ .69,  .78,  .894} 49.6 & \cellcolor[rgb]{ .667,  .765,  .886} 53.0 & \cellcolor[rgb]{ .659,  .757,  .886} 54.2 & \cellcolor[rgb]{ .702,  .788,  .898} 47.6 & \cellcolor[rgb]{ .635,  .741,  .875} 58.2 & \cellcolor[rgb]{ .698,  .784,  .898} 48.4 \\
    Deepseekcoder-V2 & 236B  & \cellcolor[rgb]{ .6,  .714,  .863} 63.8 & \cellcolor[rgb]{ .643,  .745,  .878} 57.0 & \cellcolor[rgb]{ .58,  .702,  .859} 66.6 & \cellcolor[rgb]{ .596,  .714,  .863} 64.0 & \cellcolor[rgb]{ .592,  .71,  .863} 64.8 & \cellcolor[rgb]{ .58,  .702,  .855} 67.0 & \cellcolor[rgb]{ .635,  .741,  .875} 58.4 & \cellcolor[rgb]{ .612,  .722,  .867} 62.0 & \cellcolor[rgb]{ .616,  .725,  .871} 61.4 & \cellcolor[rgb]{ .596,  .714,  .863} 64.2 & \cellcolor[rgb]{ .596,  .714,  .863} 64.0 & \cellcolor[rgb]{ .588,  .706,  .859} 65.8 & \cellcolor[rgb]{ .635,  .741,  .875} 58.0 & \cellcolor[rgb]{ .604,  .718,  .867} 63.2 & \cellcolor[rgb]{ .6,  .718,  .863} 63.6 & \cellcolor[rgb]{ .635,  .741,  .875} 58.2 & \cellcolor[rgb]{ .608,  .722,  .867} 62.4 & \cellcolor[rgb]{ .608,  .722,  .867} 62.6 & \cellcolor[rgb]{ .58,  .702,  .859} 66.6 \\
    Qwen2-Instruct & 72B   & \cellcolor[rgb]{ .675,  .769,  .89} 52.0 & \cellcolor[rgb]{ .659,  .757,  .886} 54.2 & \cellcolor[rgb]{ .69,  .78,  .894} 49.6 & \cellcolor[rgb]{ .651,  .753,  .882} 55.4 & \cellcolor[rgb]{ .686,  .776,  .894} 50.0 & \cellcolor[rgb]{ .675,  .769,  .89} 51.6 & \cellcolor[rgb]{ .678,  .773,  .89} 51.0 & \cellcolor[rgb]{ .678,  .773,  .89} 51.2 & \cellcolor[rgb]{ .702,  .788,  .898} 47.8 & \cellcolor[rgb]{ .655,  .753,  .882} 55.2 & \cellcolor[rgb]{ .671,  .765,  .886} 52.4 & \cellcolor[rgb]{ .667,  .765,  .886} 53.2 & \cellcolor[rgb]{ .702,  .788,  .898} 47.8 & \cellcolor[rgb]{ .659,  .757,  .882} 54.4 & \cellcolor[rgb]{ .639,  .745,  .878} 57.2 & \cellcolor[rgb]{ .682,  .776,  .894} 50.6 & \cellcolor[rgb]{ .671,  .765,  .886} 52.4 & \cellcolor[rgb]{ .675,  .769,  .89} 51.6 & \cellcolor[rgb]{ .675,  .769,  .89} 52.0 \\
    CodeLlama-Python & 34B   & \cellcolor[rgb]{ .757,  .827,  .918} 38.8 & \cellcolor[rgb]{ .749,  .824,  .914} 40.0 & \cellcolor[rgb]{ .757,  .827,  .918} 39.2 & \cellcolor[rgb]{ .757,  .827,  .918} 39.0 & \cellcolor[rgb]{ .741,  .816,  .914} 41.4 & \cellcolor[rgb]{ .714,  .796,  .902} 45.8 & \cellcolor[rgb]{ .718,  .8,  .906} 44.8 & \cellcolor[rgb]{ .718,  .8,  .906} 45.0 & \cellcolor[rgb]{ .729,  .808,  .91} 43.2 & \cellcolor[rgb]{ .698,  .788,  .898} 48.0 & \cellcolor[rgb]{ .706,  .792,  .902} 46.8 & \cellcolor[rgb]{ .737,  .812,  .91} 42.2 & \cellcolor[rgb]{ .757,  .827,  .918} 38.8 & \cellcolor[rgb]{ .725,  .804,  .906} 44.0 & \cellcolor[rgb]{ .722,  .804,  .906} 44.2 & \cellcolor[rgb]{ .729,  .808,  .91} 43.0 & \cellcolor[rgb]{ .722,  .8,  .906} 44.6 & \cellcolor[rgb]{ .718,  .8,  .906} 45.0 & \cellcolor[rgb]{ .725,  .804,  .906} 44.0 \\
    CodeLlama-Instruct & 34B   & \cellcolor[rgb]{ .722,  .8,  .906} 44.6 & \cellcolor[rgb]{ .698,  .784,  .898} 48.4 & \cellcolor[rgb]{ .725,  .804,  .906} 43.8 & \cellcolor[rgb]{ .714,  .796,  .902} 46.0 & \cellcolor[rgb]{ .722,  .804,  .906} 44.4 & \cellcolor[rgb]{ .671,  .765,  .886} 52.6 & \cellcolor[rgb]{ .682,  .776,  .894} 50.4 & \cellcolor[rgb]{ .69,  .78,  .894} 49.4 & \cellcolor[rgb]{ .714,  .796,  .902} 46.0 & \cellcolor[rgb]{ .675,  .769,  .89} 52.0 & \cellcolor[rgb]{ .678,  .773,  .89} 51.2 & \cellcolor[rgb]{ .698,  .784,  .898} 48.4 & \cellcolor[rgb]{ .733,  .812,  .91} 42.4 & \cellcolor[rgb]{ .698,  .784,  .898} 48.2 & \cellcolor[rgb]{ .694,  .784,  .898} 48.6 & \cellcolor[rgb]{ .698,  .788,  .898} 48.0 & \cellcolor[rgb]{ .71,  .796,  .902} 46.2 & \cellcolor[rgb]{ .69,  .78,  .894} 49.4 & \cellcolor[rgb]{ .667,  .765,  .886} 53.2 \\
    CodeLlama & 34B   & \cellcolor[rgb]{ .749,  .82,  .914} 40.4 & \cellcolor[rgb]{ .722,  .8,  .906} 44.6 & \cellcolor[rgb]{ .714,  .796,  .902} 45.6 & \cellcolor[rgb]{ .741,  .816,  .914} 41.2 & \cellcolor[rgb]{ .757,  .827,  .918} 39.0 & \cellcolor[rgb]{ .686,  .776,  .894} 50.0 & \cellcolor[rgb]{ .694,  .78,  .894} 49.0 & \cellcolor[rgb]{ .706,  .792,  .898} 47.0 & \cellcolor[rgb]{ .71,  .792,  .902} 46.6 & \cellcolor[rgb]{ .694,  .784,  .898} 48.8 & \cellcolor[rgb]{ .686,  .776,  .894} 49.8 & \cellcolor[rgb]{ .702,  .788,  .898} 47.6 & \cellcolor[rgb]{ .753,  .824,  .918} 39.8 & \cellcolor[rgb]{ .71,  .792,  .902} 46.6 & \cellcolor[rgb]{ .706,  .792,  .902} 46.8 & \cellcolor[rgb]{ .722,  .8,  .906} 44.6 & \cellcolor[rgb]{ .722,  .804,  .906} 44.4 & \cellcolor[rgb]{ .686,  .776,  .894} 50.0 & \cellcolor[rgb]{ .694,  .784,  .898} 48.6 \\
    WizardCoder-V1.1 & 33B   & \cellcolor[rgb]{ .718,  .8,  .906} 44.8 & \cellcolor[rgb]{ .843,  .89,  .949} 25.4 & \cellcolor[rgb]{ .71,  .792,  .902} 46.4 & \cellcolor[rgb]{ .702,  .788,  .898} 47.6 & \cellcolor[rgb]{ .698,  .784,  .898} 48.4 & \cellcolor[rgb]{ .714,  .796,  .902} 45.6 & \cellcolor[rgb]{ .69,  .78,  .894} 49.2 & \cellcolor[rgb]{ .694,  .784,  .898} 48.8 & \cellcolor[rgb]{ .722,  .8,  .906} 44.6 & \cellcolor[rgb]{ .686,  .776,  .894} 50.0 & \cellcolor[rgb]{ .686,  .776,  .894} 50.0 & \cellcolor[rgb]{ .718,  .8,  .906} 45.0 & \cellcolor[rgb]{ .733,  .812,  .91} 42.4 & \cellcolor[rgb]{ .69,  .78,  .894} 49.2 & \cellcolor[rgb]{ .698,  .784,  .898} 48.2 & \cellcolor[rgb]{ .698,  .784,  .898} 48.2 & \cellcolor[rgb]{ .718,  .8,  .902} 45.4 & \cellcolor[rgb]{ .678,  .773,  .89} 51.0 & \cellcolor[rgb]{ .71,  .792,  .902} 46.4 \\
    Deepseekcoder-instruct & 33B   & \cellcolor[rgb]{ .714,  .796,  .902} 46.0 & \cellcolor[rgb]{ .725,  .808,  .906} 43.6 & \cellcolor[rgb]{ .686,  .776,  .894} 49.8 & \cellcolor[rgb]{ .694,  .78,  .894} 49.0 & \cellcolor[rgb]{ .706,  .792,  .902} 46.8 & \cellcolor[rgb]{ .694,  .784,  .898} 48.8 & \cellcolor[rgb]{ .706,  .792,  .898} 47.0 & \cellcolor[rgb]{ .686,  .776,  .894} 50.0 & \cellcolor[rgb]{ .706,  .792,  .902} 46.8 & \cellcolor[rgb]{ .675,  .769,  .89} 52.0 & \cellcolor[rgb]{ .675,  .769,  .89} 51.8 & \cellcolor[rgb]{ .698,  .784,  .898} 48.2 & \cellcolor[rgb]{ .741,  .816,  .914} 41.6 & \cellcolor[rgb]{ .675,  .769,  .89} 52.0 & \cellcolor[rgb]{ .698,  .784,  .898} 48.4 & \cellcolor[rgb]{ .706,  .792,  .898} 47.0 & \cellcolor[rgb]{ .698,  .784,  .898} 48.2 & \cellcolor[rgb]{ .675,  .769,  .89} 52.2 & \cellcolor[rgb]{ .69,  .78,  .894} 49.6 \\
    Deepseekcoder-base & 33B   & \cellcolor[rgb]{ .741,  .816,  .914} 41.2 & \cellcolor[rgb]{ .733,  .812,  .91} 42.8 & \cellcolor[rgb]{ .729,  .808,  .91} 43.2 & \cellcolor[rgb]{ .714,  .796,  .902} 45.6 & \cellcolor[rgb]{ .725,  .804,  .906} 43.8 & \cellcolor[rgb]{ .714,  .796,  .902} 46.0 & \cellcolor[rgb]{ .702,  .788,  .898} 47.6 & \cellcolor[rgb]{ .702,  .788,  .898} 47.4 & \cellcolor[rgb]{ .706,  .788,  .898} 47.2 & \cellcolor[rgb]{ .694,  .784,  .898} 48.6 & \cellcolor[rgb]{ .69,  .78,  .894} 49.2 & \cellcolor[rgb]{ .682,  .776,  .894} 50.6 & \cellcolor[rgb]{ .733,  .812,  .91} 42.8 & \cellcolor[rgb]{ .702,  .788,  .898} 47.4 & \cellcolor[rgb]{ .706,  .792,  .902} 46.8 & \cellcolor[rgb]{ .725,  .804,  .906} 44.0 & \cellcolor[rgb]{ .71,  .792,  .902} 46.4 & \cellcolor[rgb]{ .698,  .784,  .898} 48.2 & \cellcolor[rgb]{ .718,  .8,  .906} 45.0 \\
    Starcoder2 & 15B   & \cellcolor[rgb]{ .741,  .816,  .914} 41.4 & \cellcolor[rgb]{ .725,  .804,  .906} 43.8 & \cellcolor[rgb]{ .675,  .769,  .89} 51.6 & \cellcolor[rgb]{ .718,  .8,  .906} 45.2 & \cellcolor[rgb]{ .733,  .812,  .91} 42.6 & \cellcolor[rgb]{ .725,  .804,  .906} 44.0 & \cellcolor[rgb]{ .698,  .784,  .898} 48.2 & \cellcolor[rgb]{ .722,  .8,  .906} 44.6 & \cellcolor[rgb]{ .718,  .8,  .906} 44.8 & \cellcolor[rgb]{ .686,  .776,  .894} 49.8 & \cellcolor[rgb]{ .71,  .792,  .902} 46.6 & \cellcolor[rgb]{ .714,  .796,  .902} 45.8 & \cellcolor[rgb]{ .718,  .8,  .906} 45.0 & \cellcolor[rgb]{ .694,  .78,  .894} 49.0 & \cellcolor[rgb]{ .71,  .792,  .902} 46.6 & \cellcolor[rgb]{ .769,  .835,  .922} 37.0 & \cellcolor[rgb]{ .702,  .788,  .898} 47.4 & \cellcolor[rgb]{ .675,  .769,  .89} 52.2 & \cellcolor[rgb]{ .71,  .796,  .902} 46.2 \\
    WizardCoder-V1.0 & 15B   & \cellcolor[rgb]{ .82,  .871,  .937} 29.2 & \cellcolor[rgb]{ .812,  .867,  .937} 30.0 & \cellcolor[rgb]{ .808,  .867,  .937} 30.6 & \cellcolor[rgb]{ .824,  .875,  .941} 28.6 & \cellcolor[rgb]{ .816,  .871,  .937} 29.6 & \cellcolor[rgb]{ .796,  .855,  .929} 33.0 & \cellcolor[rgb]{ .784,  .847,  .925} 34.8 & \cellcolor[rgb]{ .792,  .851,  .929} 33.6 & \cellcolor[rgb]{ .773,  .839,  .922} 36.2 & \cellcolor[rgb]{ .769,  .835,  .922} 36.8 & \cellcolor[rgb]{ .792,  .855,  .929} 33.2 & \cellcolor[rgb]{ .792,  .851,  .929} 33.4 & \cellcolor[rgb]{ .773,  .839,  .922} 36.4 & \cellcolor[rgb]{ .792,  .851,  .929} 33.6 & \cellcolor[rgb]{ .796,  .855,  .929} 33.0 & \cellcolor[rgb]{ .82,  .871,  .937} 29.0 & \cellcolor[rgb]{ .78,  .843,  .925} 35.0 & \cellcolor[rgb]{ .788,  .851,  .929} 34.0 & \cellcolor[rgb]{ .796,  .855,  .933} 32.4 \\
    Starcoder & 15B   & \cellcolor[rgb]{ .824,  .875,  .941} 28.2 & \cellcolor[rgb]{ .812,  .867,  .937} 30.0 & \cellcolor[rgb]{ .796,  .855,  .929} 33.0 & \cellcolor[rgb]{ .792,  .855,  .929} 33.2 & \cellcolor[rgb]{ .792,  .851,  .929} 33.4 & \cellcolor[rgb]{ .78,  .843,  .925} 35.2 & \cellcolor[rgb]{ .784,  .847,  .929} 34.4 & \cellcolor[rgb]{ .804,  .859,  .933} 31.6 & \cellcolor[rgb]{ .788,  .851,  .929} 34.0 & \cellcolor[rgb]{ .773,  .839,  .922} 36.4 & \cellcolor[rgb]{ .784,  .847,  .925} 34.8 & \cellcolor[rgb]{ .792,  .851,  .929} 33.4 & \cellcolor[rgb]{ .773,  .839,  .922} 36.6 & \cellcolor[rgb]{ .78,  .843,  .925} 35.0 & \cellcolor[rgb]{ .784,  .847,  .925} 34.8 & \cellcolor[rgb]{ .831,  .878,  .941} 27.4 & \cellcolor[rgb]{ .769,  .835,  .922} 37.0 & \cellcolor[rgb]{ .808,  .863,  .933} 30.8 & \cellcolor[rgb]{ .792,  .855,  .929} 33.2 \\
    phi-3-instruct & 14B   & \cellcolor[rgb]{ .8,  .859,  .933} 31.8 & \cellcolor[rgb]{ .839,  .886,  .945} 26.0 & \cellcolor[rgb]{ .757,  .827,  .918} 38.8 & \cellcolor[rgb]{ .773,  .839,  .922} 36.4 & \cellcolor[rgb]{ .769,  .835,  .922} 37.2 & \cellcolor[rgb]{ .733,  .812,  .91} 42.4 & \cellcolor[rgb]{ .773,  .839,  .922} 36.2 & \cellcolor[rgb]{ .769,  .835,  .922} 37.2 & \cellcolor[rgb]{ .776,  .843,  .925} 35.6 & \cellcolor[rgb]{ .741,  .816,  .914} 41.2 & \cellcolor[rgb]{ .729,  .808,  .91} 43.4 & \cellcolor[rgb]{ .757,  .827,  .918} 39.2 & \cellcolor[rgb]{ .847,  .894,  .949} 24.4 & \cellcolor[rgb]{ .776,  .839,  .925} 36.0 & \cellcolor[rgb]{ .769,  .835,  .922} 36.8 & \cellcolor[rgb]{ .761,  .831,  .922} 38.0 & \cellcolor[rgb]{ .792,  .851,  .929} 33.6 & \cellcolor[rgb]{ .741,  .816,  .914} 41.2 & \cellcolor[rgb]{ .733,  .812,  .91} 42.8 \\
    Llama-3-Instruct & 8B    & \cellcolor[rgb]{ .769,  .835,  .922} 37.0 & \cellcolor[rgb]{ .773,  .839,  .922} 36.4 & \cellcolor[rgb]{ .78,  .843,  .925} 35.0 & \cellcolor[rgb]{ .757,  .827,  .918} 38.6 & \cellcolor[rgb]{ .773,  .839,  .922} 36.2 & \cellcolor[rgb]{ .761,  .831,  .918} 38.4 & \cellcolor[rgb]{ .753,  .824,  .918} 39.6 & \cellcolor[rgb]{ .749,  .824,  .914} 40.0 & \cellcolor[rgb]{ .773,  .839,  .922} 36.2 & \cellcolor[rgb]{ .773,  .839,  .922} 36.6 & \cellcolor[rgb]{ .761,  .831,  .918} 38.4 & \cellcolor[rgb]{ .737,  .812,  .91} 42.2 & \cellcolor[rgb]{ .851,  .894,  .949} 24.2 & \cellcolor[rgb]{ .776,  .843,  .925} 35.8 & \cellcolor[rgb]{ .765,  .831,  .922} 37.6 & \cellcolor[rgb]{ .761,  .831,  .922} 38.0 & \cellcolor[rgb]{ .804,  .859,  .933} 31.6 & \cellcolor[rgb]{ .737,  .812,  .91} 42.2 & \cellcolor[rgb]{ .761,  .831,  .918} 38.2 \\
    CodeQwen1.5-Chat & 7B    & \cellcolor[rgb]{ .733,  .812,  .91} 42.8 & \cellcolor[rgb]{ .737,  .812,  .91} 42.0 & \cellcolor[rgb]{ .729,  .808,  .91} 43.0 & \cellcolor[rgb]{ .71,  .792,  .902} 46.4 & \cellcolor[rgb]{ .722,  .8,  .906} 44.6 & \cellcolor[rgb]{ .725,  .804,  .906} 43.8 & \cellcolor[rgb]{ .737,  .812,  .91} 42.2 & \cellcolor[rgb]{ .733,  .812,  .91} 42.8 & \cellcolor[rgb]{ .741,  .816,  .914} 41.6 & \cellcolor[rgb]{ .718,  .8,  .906} 44.8 & \cellcolor[rgb]{ .729,  .808,  .91} 43.0 & \cellcolor[rgb]{ .729,  .808,  .91} 43.4 & \cellcolor[rgb]{ .761,  .831,  .918} 38.2 & \cellcolor[rgb]{ .725,  .808,  .906} 43.6 & \cellcolor[rgb]{ .737,  .812,  .91} 42.0 & \cellcolor[rgb]{ .753,  .824,  .918} 39.4 & \cellcolor[rgb]{ .71,  .792,  .902} 46.6 & \cellcolor[rgb]{ .714,  .796,  .902} 45.8 & \cellcolor[rgb]{ .725,  .808,  .906} 43.6 \\
    CodeLlama-Instruct & 7B    & \cellcolor[rgb]{ .757,  .827,  .918} 38.6 & \cellcolor[rgb]{ .776,  .839,  .925} 36.0 & \cellcolor[rgb]{ .761,  .831,  .918} 38.4 & \cellcolor[rgb]{ .761,  .831,  .918} 38.4 & \cellcolor[rgb]{ .761,  .831,  .918} 38.2 & \cellcolor[rgb]{ .753,  .824,  .918} 39.6 & \cellcolor[rgb]{ .737,  .812,  .91} 42.2 & \cellcolor[rgb]{ .729,  .808,  .91} 43.4 & \cellcolor[rgb]{ .773,  .839,  .922} 36.4 & \cellcolor[rgb]{ .749,  .82,  .914} 40.4 & \cellcolor[rgb]{ .745,  .82,  .914} 41.0 & \cellcolor[rgb]{ .745,  .82,  .914} 41.0 & \cellcolor[rgb]{ .757,  .827,  .918} 38.8 & \cellcolor[rgb]{ .741,  .816,  .914} 41.6 & \cellcolor[rgb]{ .765,  .831,  .922} 37.6 & \cellcolor[rgb]{ .733,  .812,  .91} 42.6 & \cellcolor[rgb]{ .753,  .824,  .918} 39.6 & \cellcolor[rgb]{ .749,  .824,  .914} 40.2 & \cellcolor[rgb]{ .745,  .82,  .914} 41.0 \\
    CodeLlama-hf & 7B    & \cellcolor[rgb]{ .773,  .839,  .922} 36.4 & \cellcolor[rgb]{ .773,  .839,  .922} 36.2 & \cellcolor[rgb]{ .769,  .835,  .922} 36.8 & \cellcolor[rgb]{ .784,  .847,  .925} 34.6 & \cellcolor[rgb]{ .773,  .839,  .922} 36.4 & \cellcolor[rgb]{ .773,  .839,  .922} 36.6 & \cellcolor[rgb]{ .749,  .824,  .914} 40.2 & \cellcolor[rgb]{ .753,  .824,  .918} 39.6 & \cellcolor[rgb]{ .776,  .839,  .925} 36.0 & \cellcolor[rgb]{ .753,  .824,  .918} 39.4 & \cellcolor[rgb]{ .749,  .824,  .914} 40.2 & \cellcolor[rgb]{ .749,  .824,  .914} 40.0 & \cellcolor[rgb]{ .773,  .839,  .922} 36.6 & \cellcolor[rgb]{ .757,  .827,  .918} 39.2 & \cellcolor[rgb]{ .78,  .843,  .925} 35.4 & \cellcolor[rgb]{ .765,  .831,  .922} 37.8 & \cellcolor[rgb]{ .769,  .835,  .922} 36.8 & \cellcolor[rgb]{ .757,  .827,  .918} 39.2 & \cellcolor[rgb]{ .757,  .827,  .918} 38.8 \\
    Deepseekcoder-instruct & 6.7B  & \cellcolor[rgb]{ .78,  .843,  .925} 35.0 & \cellcolor[rgb]{ .769,  .835,  .922} 37.0 & \cellcolor[rgb]{ .776,  .843,  .925} 35.6 & \cellcolor[rgb]{ .749,  .82,  .914} 40.4 & \cellcolor[rgb]{ .78,  .843,  .925} 35.0 & \cellcolor[rgb]{ .773,  .839,  .922} 36.6 & \cellcolor[rgb]{ .757,  .827,  .918} 39.2 & \cellcolor[rgb]{ .757,  .827,  .918} 38.8 & \cellcolor[rgb]{ .753,  .824,  .918} 39.4 & \cellcolor[rgb]{ .737,  .812,  .91} 42.2 & \cellcolor[rgb]{ .761,  .831,  .918} 38.2 & \cellcolor[rgb]{ .737,  .812,  .91} 42.0 & \cellcolor[rgb]{ .769,  .835,  .922} 37.2 & \cellcolor[rgb]{ .749,  .824,  .914} 40.2 & \cellcolor[rgb]{ .765,  .835,  .922} 37.4 & \cellcolor[rgb]{ .769,  .835,  .922} 36.8 & \cellcolor[rgb]{ .733,  .812,  .91} 42.8 & \cellcolor[rgb]{ .745,  .82,  .914} 40.8 & \cellcolor[rgb]{ .788,  .847,  .929} 34.2 \\
    Deepseekcoder-base & 6.7B  & \cellcolor[rgb]{ .757,  .827,  .918} 38.8 & \cellcolor[rgb]{ .733,  .812,  .91} 42.4 & \cellcolor[rgb]{ .741,  .816,  .914} 41.2 & \cellcolor[rgb]{ .729,  .808,  .91} 43.2 & \cellcolor[rgb]{ .749,  .82,  .914} 40.4 & \cellcolor[rgb]{ .725,  .808,  .906} 43.6 & \cellcolor[rgb]{ .733,  .812,  .91} 42.6 & \cellcolor[rgb]{ .733,  .812,  .91} 42.8 & \cellcolor[rgb]{ .741,  .816,  .914} 41.6 & \cellcolor[rgb]{ .71,  .792,  .902} 46.4 & \cellcolor[rgb]{ .741,  .816,  .914} 41.4 & \cellcolor[rgb]{ .71,  .796,  .902} 46.2 & \cellcolor[rgb]{ .729,  .808,  .91} 43.0 & \cellcolor[rgb]{ .722,  .8,  .906} 44.6 & \cellcolor[rgb]{ .741,  .816,  .914} 41.6 & \cellcolor[rgb]{ .745,  .82,  .914} 40.8 & \cellcolor[rgb]{ .718,  .8,  .906} 44.8 & \cellcolor[rgb]{ .729,  .808,  .91} 43.4 & \cellcolor[rgb]{ .737,  .816,  .91} 41.8 \\
    CodeGen-multi & 6B    & \cellcolor[rgb]{ .82,  .875,  .941} 28.8 & \cellcolor[rgb]{ .843,  .89,  .949} 25.4 & \cellcolor[rgb]{ .965,  .976,  .988} 6.2 & \cellcolor[rgb]{ .839,  .886,  .945} 25.6 & \cellcolor[rgb]{ .773,  .839,  .922} 36.2 & \cellcolor[rgb]{ .843,  .89,  .949} 25.2 & \cellcolor[rgb]{ .894,  .925,  .965} 17.4 & \cellcolor[rgb]{ .847,  .894,  .949} 24.4 & \cellcolor[rgb]{ .761,  .831,  .918} 38.4 & \cellcolor[rgb]{ .859,  .902,  .953} 22.8 & \cellcolor[rgb]{ .859,  .902,  .953} 22.6 & \cellcolor[rgb]{ .831,  .878,  .945} 27.2 & \cellcolor[rgb]{ .902,  .929,  .969} 16.2 & \cellcolor[rgb]{ .961,  .973,  .988} 6.4 & \cellcolor[rgb]{ .882,  .918,  .961} 18.8 & \cellcolor[rgb]{ .808,  .863,  .933} 31.0 & \cellcolor[rgb]{ .694,  .784,  .898} 48.6 & \cellcolor[rgb]{ .796,  .855,  .933} 32.4 & \cellcolor[rgb]{ .843,  .89,  .949} 25.2 \\
    phi-1\_5 & 1.3B  & \cellcolor[rgb]{ .82,  .871,  .937} 29.2 & \cellcolor[rgb]{ .902,  .929,  .969} 16.0 & \cellcolor[rgb]{ .922,  .945,  .973} 13.2 & \cellcolor[rgb]{ .839,  .886,  .945} 25.8 & \cellcolor[rgb]{ .835,  .882,  .945} 26.8 & \cellcolor[rgb]{ .941,  .957,  .98} 9.8 & \cellcolor[rgb]{ .812,  .867,  .937} 30.4 & \cellcolor[rgb]{ .835,  .882,  .945} 26.6 & \cellcolor[rgb]{ .89,  .922,  .965} 17.8 & \cellcolor[rgb]{ .835,  .882,  .945} 26.6 & \cellcolor[rgb]{ .839,  .886,  .945} 25.8 & \cellcolor[rgb]{ .949,  .965,  .984} 8.4 & \cellcolor[rgb]{ .961,  .973,  .988} 6.6 & \cellcolor[rgb]{ .996,  .996,  1} 1.4 & \cellcolor[rgb]{ .843,  .89,  .949} 25.2 & \cellcolor[rgb]{ .812,  .867,  .937} 30.4 & \cellcolor[rgb]{ .784,  .847,  .929} 34.4 & \cellcolor[rgb]{ .835,  .882,  .945} 26.6 & \cellcolor[rgb]{ .808,  .863,  .933} 30.8 \\
    phi-1 & 1.3B  & 0.2   & \cellcolor[rgb]{ .961,  .973,  .988} 7.0 & \cellcolor[rgb]{ .941,  .961,  .98} 9.6 & \cellcolor[rgb]{ .98,  .988,  .996} 3.6 & \cellcolor[rgb]{ .984,  .992,  .996} 2.8 & \cellcolor[rgb]{ .894,  .925,  .965} 17.0 & \cellcolor[rgb]{ .882,  .918,  .961} 19.0 & \cellcolor[rgb]{ .894,  .925,  .965} 17.4 & \cellcolor[rgb]{ .855,  .898,  .949} 23.6 & \cellcolor[rgb]{ .945,  .961,  .98} 9.2 & \cellcolor[rgb]{ .929,  .949,  .976} 11.8 & \cellcolor[rgb]{ .945,  .961,  .98} 9.4 & \cellcolor[rgb]{ .933,  .953,  .976} 11.2 & \cellcolor[rgb]{ .961,  .973,  .988} 6.8 & \cellcolor[rgb]{ .969,  .98,  .992} 5.4 & \cellcolor[rgb]{ .992,  .996,  1} 1.8 & \cellcolor[rgb]{ .878,  .914,  .961} 19.8 & \cellcolor[rgb]{ .914,  .941,  .973} 14.0 & \cellcolor[rgb]{ .914,  .941,  .973} 14.0 \\
    \midrule
    Average &       & \cellcolor[rgb]{ 1,  .937,  .737} 40.4 & \cellcolor[rgb]{ 1,  .949,  .792} 38.3 & \cellcolor[rgb]{ 1,  .933,  .725} 40.8 & \cellcolor[rgb]{ 1,  .922,  .682} 42.4 & \cellcolor[rgb]{ 1,  .925,  .702} 41.7 & \cellcolor[rgb]{ 1,  .918,  .663} 43.1 & \cellcolor[rgb]{ 1,  .918,  .663} 43.1 & \cellcolor[rgb]{ 1,  .914,  .643} 43.9 & \cellcolor[rgb]{ 1,  .922,  .678} 42.5 & \cellcolor[rgb]{ 1,  .906,  .612} 45.0 & \cellcolor[rgb]{ 1,  .914,  .639} 44.1 & \cellcolor[rgb]{ 1,  .918,  .663} 43.2 & \cellcolor[rgb]{ 1,  .949,  .8} 38.0 & \cellcolor[rgb]{ 1,  .925,  .702} 41.7 & \cellcolor[rgb]{ 1,  .922,  .675} 42.7 & \cellcolor[rgb]{ 1,  .929,  .718} 41.1 & \cellcolor[rgb]{ 1,  .91,  .631} 44.3 & \cellcolor[rgb]{ 1,  .902,  .6} 45.4 & \cellcolor[rgb]{ 1,  .914,  .643} 43.8 \\
    \midrule
    \multicolumn{21}{c}{\textbf{Output Reasoning Performence}} \\
    \midrule
    \textbf{Models} & \textbf{Size} & \multicolumn{1}{c}{\textbf{cs}} & \multicolumn{1}{c}{\textbf{cpp}} & \multicolumn{1}{c}{\textbf{d}} & \multicolumn{1}{c}{\textbf{go}} & \multicolumn{1}{c}{\textbf{java}} & \multicolumn{1}{c}{\textbf{js}} & \multicolumn{1}{c}{\textbf{jl}} & \multicolumn{1}{c}{\textbf{lua}} & \multicolumn{1}{c}{\textbf{pl}} & \multicolumn{1}{c}{\textbf{php}} & \multicolumn{1}{c}{\textbf{py}} & \multicolumn{1}{c}{\textbf{r}} & \multicolumn{1}{c}{\textbf{rkt}} & \multicolumn{1}{c}{\textbf{rb}} & \multicolumn{1}{c}{\textbf{rs}} & \multicolumn{1}{c}{\textbf{scala}} & \multicolumn{1}{c}{\textbf{sh}} & \multicolumn{1}{c}{\textbf{swift}} & \multicolumn{1}{c}{\textbf{ts}} \\
    \midrule
    GPT-4o & -     & \cellcolor[rgb]{ .529,  .663,  .839} 75.0 & \cellcolor[rgb]{ .529,  .667,  .839} 74.8 & \cellcolor[rgb]{ .553,  .682,  .847} 71.4 & \cellcolor[rgb]{ .514,  .655,  .835} 77.0 & \cellcolor[rgb]{ .541,  .675,  .843} 73.2 & \cellcolor[rgb]{ .51,  .651,  .831} 77.6 & \cellcolor[rgb]{ .537,  .671,  .843} 73.6 & \cellcolor[rgb]{ .529,  .667,  .839} 74.8 & \cellcolor[rgb]{ .533,  .671,  .839} 74.0 & \cellcolor[rgb]{ .525,  .663,  .839} 75.4 & \cellcolor[rgb]{ .525,  .663,  .839} 75.4 & \cellcolor[rgb]{ .549,  .678,  .847} 72.0 & \cellcolor[rgb]{ .553,  .682,  .847} 70.8 & \cellcolor[rgb]{ .533,  .671,  .839} 74.0 & \cellcolor[rgb]{ .533,  .667,  .839} 74.4 & \cellcolor[rgb]{ .549,  .678,  .847} 71.8 & \cellcolor[rgb]{ .549,  .678,  .847} 71.6 & \cellcolor[rgb]{ .522,  .659,  .835} 76.0 & \cellcolor[rgb]{ .518,  .659,  .835} 76.4 \\
    GPT-4o-mini & -     & \cellcolor[rgb]{ .604,  .718,  .867} 63.0 & \cellcolor[rgb]{ .604,  .718,  .867} 63.0 & \cellcolor[rgb]{ .616,  .725,  .871} 61.4 & \cellcolor[rgb]{ .6,  .718,  .863} 63.4 & \cellcolor[rgb]{ .663,  .761,  .886} 54.0 & \cellcolor[rgb]{ .612,  .725,  .867} 61.8 & \cellcolor[rgb]{ .635,  .741,  .878} 57.8 & \cellcolor[rgb]{ .624,  .733,  .871} 60.0 & \cellcolor[rgb]{ .639,  .745,  .878} 57.4 & \cellcolor[rgb]{ .596,  .714,  .863} 64.2 & \cellcolor[rgb]{ .612,  .725,  .867} 61.6 & \cellcolor[rgb]{ .627,  .733,  .875} 59.6 & \cellcolor[rgb]{ .643,  .749,  .878} 56.6 & \cellcolor[rgb]{ .616,  .725,  .871} 61.2 & \cellcolor[rgb]{ .612,  .725,  .867} 61.8 & \cellcolor[rgb]{ .616,  .725,  .871} 61.2 & \cellcolor[rgb]{ .647,  .749,  .878} 56.2 & \cellcolor[rgb]{ .604,  .718,  .867} 63.0 & \cellcolor[rgb]{ .616,  .725,  .871} 61.2 \\
    GPT-3.5 Turbo & -     & \cellcolor[rgb]{ .659,  .757,  .886} 54.2 & \cellcolor[rgb]{ .729,  .808,  .91} 43.2 & \cellcolor[rgb]{ .647,  .749,  .882} 56.0 & \cellcolor[rgb]{ .667,  .765,  .886} 53.2 & \cellcolor[rgb]{ .725,  .808,  .906} 43.6 & \cellcolor[rgb]{ .647,  .749,  .878} 56.2 & \cellcolor[rgb]{ .659,  .757,  .886} 54.2 & \cellcolor[rgb]{ .659,  .757,  .882} 54.6 & \cellcolor[rgb]{ .675,  .769,  .89} 51.8 & \cellcolor[rgb]{ .655,  .753,  .882} 55.2 & \cellcolor[rgb]{ .639,  .745,  .878} 57.2 & \cellcolor[rgb]{ .69,  .78,  .894} 49.4 & \cellcolor[rgb]{ .698,  .788,  .898} 48.0 & \cellcolor[rgb]{ .647,  .749,  .878} 56.4 & \cellcolor[rgb]{ .659,  .757,  .882} 54.6 & \cellcolor[rgb]{ .647,  .749,  .878} 56.4 & \cellcolor[rgb]{ .678,  .773,  .89} 51.0 & \cellcolor[rgb]{ .635,  .741,  .878} 57.8 & \cellcolor[rgb]{ .663,  .761,  .886} 53.6 \\
    Deepseekcoder-V2 & 236B  & \cellcolor[rgb]{ .58,  .702,  .859} 66.6 & \cellcolor[rgb]{ .584,  .706,  .859} 66.2 & \cellcolor[rgb]{ .6,  .718,  .863} 63.4 & \cellcolor[rgb]{ .573,  .698,  .855} 68.0 & \cellcolor[rgb]{ .576,  .698,  .855} 67.6 & \cellcolor[rgb]{ .588,  .71,  .859} 65.4 & \cellcolor[rgb]{ .592,  .71,  .863} 64.8 & \cellcolor[rgb]{ .6,  .718,  .863} 63.6 & \cellcolor[rgb]{ .604,  .718,  .867} 63.0 & \cellcolor[rgb]{ .576,  .698,  .855} 67.4 & \cellcolor[rgb]{ .58,  .702,  .859} 66.8 & \cellcolor[rgb]{ .604,  .718,  .867} 63.0 & \cellcolor[rgb]{ .608,  .722,  .867} 62.2 & \cellcolor[rgb]{ .592,  .71,  .859} 65.2 & \cellcolor[rgb]{ .588,  .706,  .859} 65.8 & \cellcolor[rgb]{ .604,  .718,  .867} 63.2 & \cellcolor[rgb]{ .631,  .737,  .875} 58.8 & \cellcolor[rgb]{ .573,  .698,  .855} 67.8 & \cellcolor[rgb]{ .584,  .702,  .859} 66.4 \\
    Qwen2-Instruct & 72B   & \cellcolor[rgb]{ .678,  .773,  .89} 51.2 & \cellcolor[rgb]{ .686,  .776,  .894} 50.2 & \cellcolor[rgb]{ .675,  .769,  .89} 51.6 & \cellcolor[rgb]{ .663,  .761,  .886} 53.6 & \cellcolor[rgb]{ .761,  .831,  .918} 38.2 & \cellcolor[rgb]{ .675,  .769,  .89} 52.0 & \cellcolor[rgb]{ .678,  .773,  .89} 51.0 & \cellcolor[rgb]{ .694,  .78,  .894} 49.0 & \cellcolor[rgb]{ .714,  .796,  .902} 45.8 & \cellcolor[rgb]{ .682,  .773,  .89} 50.8 & \cellcolor[rgb]{ .678,  .773,  .89} 51.2 & \cellcolor[rgb]{ .718,  .8,  .906} 45.0 & \cellcolor[rgb]{ .706,  .792,  .902} 46.8 & \cellcolor[rgb]{ .682,  .773,  .89} 50.8 & \cellcolor[rgb]{ .678,  .773,  .89} 51.0 & \cellcolor[rgb]{ .678,  .773,  .89} 51.0 & \cellcolor[rgb]{ .714,  .796,  .902} 45.6 & \cellcolor[rgb]{ .682,  .776,  .894} 50.4 & \cellcolor[rgb]{ .667,  .765,  .886} 53.2 \\
    CodeLlama-Python & 34B   & \cellcolor[rgb]{ .741,  .816,  .914} 41.4 & \cellcolor[rgb]{ .718,  .8,  .906} 44.8 & \cellcolor[rgb]{ .714,  .796,  .902} 45.6 & \cellcolor[rgb]{ .737,  .816,  .91} 41.8 & \cellcolor[rgb]{ .741,  .816,  .914} 41.4 & \cellcolor[rgb]{ .718,  .8,  .902} 45.4 & \cellcolor[rgb]{ .718,  .8,  .906} 45.2 & \cellcolor[rgb]{ .733,  .812,  .91} 42.8 & \cellcolor[rgb]{ .725,  .808,  .906} 43.6 & \cellcolor[rgb]{ .725,  .804,  .906} 43.8 & \cellcolor[rgb]{ .725,  .804,  .906} 43.8 & \cellcolor[rgb]{ .733,  .812,  .91} 42.4 & \cellcolor[rgb]{ .757,  .827,  .918} 38.6 & \cellcolor[rgb]{ .733,  .812,  .91} 42.8 & \cellcolor[rgb]{ .71,  .792,  .902} 46.6 & \cellcolor[rgb]{ .725,  .804,  .906} 43.8 & \cellcolor[rgb]{ .737,  .812,  .91} 42.0 & \cellcolor[rgb]{ .722,  .804,  .906} 44.4 & \cellcolor[rgb]{ .718,  .8,  .906} 44.8 \\
    CodeLlama-Instruct & 34B   & \cellcolor[rgb]{ .722,  .804,  .906} 44.4 & \cellcolor[rgb]{ .71,  .796,  .902} 46.2 & \cellcolor[rgb]{ .714,  .796,  .902} 45.8 & \cellcolor[rgb]{ .706,  .792,  .902} 46.8 & \cellcolor[rgb]{ .745,  .82,  .914} 40.6 & \cellcolor[rgb]{ .702,  .788,  .898} 47.4 & \cellcolor[rgb]{ .714,  .796,  .902} 45.6 & \cellcolor[rgb]{ .733,  .812,  .91} 42.8 & \cellcolor[rgb]{ .725,  .804,  .906} 44.0 & \cellcolor[rgb]{ .718,  .8,  .906} 44.8 & \cellcolor[rgb]{ .725,  .804,  .906} 44.0 & \cellcolor[rgb]{ .749,  .824,  .914} 40.2 & \cellcolor[rgb]{ .761,  .831,  .918} 38.2 & \cellcolor[rgb]{ .722,  .804,  .906} 44.2 & \cellcolor[rgb]{ .71,  .792,  .902} 46.4 & \cellcolor[rgb]{ .725,  .804,  .906} 43.8 & \cellcolor[rgb]{ .745,  .82,  .914} 40.6 & \cellcolor[rgb]{ .718,  .8,  .906} 45.2 & \cellcolor[rgb]{ .718,  .8,  .906} 45.0 \\
    CodeLlama & 34B   & \cellcolor[rgb]{ .722,  .8,  .906} 44.6 & \cellcolor[rgb]{ .702,  .788,  .898} 47.8 & \cellcolor[rgb]{ .722,  .804,  .906} 44.2 & \cellcolor[rgb]{ .718,  .8,  .906} 45.2 & \cellcolor[rgb]{ .761,  .831,  .918} 38.4 & \cellcolor[rgb]{ .706,  .792,  .898} 47.0 & \cellcolor[rgb]{ .714,  .796,  .902} 45.8 & \cellcolor[rgb]{ .733,  .812,  .91} 42.8 & \cellcolor[rgb]{ .725,  .804,  .906} 43.8 & \cellcolor[rgb]{ .71,  .792,  .902} 46.4 & \cellcolor[rgb]{ .71,  .792,  .902} 46.4 & \cellcolor[rgb]{ .757,  .827,  .918} 38.8 & \cellcolor[rgb]{ .761,  .831,  .918} 38.4 & \cellcolor[rgb]{ .718,  .8,  .902} 45.4 & \cellcolor[rgb]{ .706,  .788,  .898} 47.2 & \cellcolor[rgb]{ .702,  .788,  .898} 47.4 & \cellcolor[rgb]{ .725,  .804,  .906} 43.8 & \cellcolor[rgb]{ .702,  .788,  .898} 47.6 & \cellcolor[rgb]{ .702,  .788,  .898} 47.4 \\
    WizardCoder-V1.1 & 33B   & \cellcolor[rgb]{ .706,  .792,  .898} 47.0 & \cellcolor[rgb]{ .706,  .792,  .902} 46.8 & \cellcolor[rgb]{ .714,  .796,  .902} 45.8 & \cellcolor[rgb]{ .722,  .804,  .906} 44.2 & \cellcolor[rgb]{ .682,  .773,  .89} 50.8 & \cellcolor[rgb]{ .686,  .776,  .894} 50.0 & \cellcolor[rgb]{ .706,  .792,  .898} 47.0 & \cellcolor[rgb]{ .714,  .796,  .902} 46.0 & \cellcolor[rgb]{ .718,  .8,  .906} 45.2 & \cellcolor[rgb]{ .678,  .773,  .89} 51.4 & \cellcolor[rgb]{ .69,  .78,  .894} 49.6 & \cellcolor[rgb]{ .725,  .804,  .906} 44.0 & \cellcolor[rgb]{ .733,  .812,  .91} 42.4 & \cellcolor[rgb]{ .698,  .784,  .898} 48.2 & \cellcolor[rgb]{ .702,  .788,  .898} 47.8 & \cellcolor[rgb]{ .718,  .8,  .906} 45.0 & \cellcolor[rgb]{ .722,  .804,  .906} 44.4 & \cellcolor[rgb]{ .698,  .788,  .898} 48.0 & \cellcolor[rgb]{ .686,  .776,  .894} 49.8 \\
    Deepseekcoder-instruct & 33B   & \cellcolor[rgb]{ .675,  .769,  .89} 52.0 & \cellcolor[rgb]{ .678,  .773,  .89} 51.4 & \cellcolor[rgb]{ .694,  .78,  .894} 49.0 & \cellcolor[rgb]{ .694,  .784,  .898} 48.8 & \cellcolor[rgb]{ .667,  .765,  .886} 53.2 & \cellcolor[rgb]{ .655,  .753,  .882} 55.0 & \cellcolor[rgb]{ .682,  .776,  .894} 50.4 & \cellcolor[rgb]{ .682,  .776,  .894} 50.4 & \cellcolor[rgb]{ .686,  .776,  .894} 50.0 & \cellcolor[rgb]{ .667,  .765,  .886} 53.0 & \cellcolor[rgb]{ .675,  .769,  .89} 52.2 & \cellcolor[rgb]{ .698,  .784,  .898} 48.2 & \cellcolor[rgb]{ .71,  .792,  .902} 46.6 & \cellcolor[rgb]{ .671,  .765,  .886} 52.8 & \cellcolor[rgb]{ .682,  .776,  .894} 50.6 & \cellcolor[rgb]{ .698,  .788,  .898} 48.0 & \cellcolor[rgb]{ .69,  .78,  .894} 49.4 & \cellcolor[rgb]{ .671,  .765,  .886} 52.8 & \cellcolor[rgb]{ .663,  .761,  .886} 53.6 \\
    DeepseekCoder-base & 33B   & \cellcolor[rgb]{ .698,  .784,  .898} 48.2 & \cellcolor[rgb]{ .686,  .776,  .894} 50.0 & \cellcolor[rgb]{ .714,  .796,  .902} 46.0 & \cellcolor[rgb]{ .694,  .784,  .898} 48.6 & \cellcolor[rgb]{ .69,  .78,  .894} 49.2 & \cellcolor[rgb]{ .678,  .773,  .89} 51.4 & \cellcolor[rgb]{ .706,  .792,  .902} 46.8 & \cellcolor[rgb]{ .698,  .788,  .898} 48.0 & \cellcolor[rgb]{ .698,  .784,  .898} 48.4 & \cellcolor[rgb]{ .675,  .769,  .89} 52.0 & \cellcolor[rgb]{ .686,  .776,  .894} 49.8 & \cellcolor[rgb]{ .718,  .8,  .906} 45.2 & \cellcolor[rgb]{ .71,  .792,  .902} 46.4 & \cellcolor[rgb]{ .694,  .78,  .894} 49.0 & \cellcolor[rgb]{ .71,  .796,  .902} 46.2 & \cellcolor[rgb]{ .702,  .788,  .898} 47.6 & \cellcolor[rgb]{ .714,  .796,  .902} 46.0 & \cellcolor[rgb]{ .69,  .78,  .894} 49.2 & \cellcolor[rgb]{ .678,  .773,  .89} 51.2 \\
    Starcoder2 & 15B   & \cellcolor[rgb]{ .714,  .796,  .902} 46.0 & \cellcolor[rgb]{ .702,  .788,  .898} 47.4 & \cellcolor[rgb]{ .706,  .788,  .898} 47.2 & \cellcolor[rgb]{ .694,  .78,  .894} 49.0 & \cellcolor[rgb]{ .698,  .784,  .898} 48.4 & \cellcolor[rgb]{ .686,  .776,  .894} 50.0 & \cellcolor[rgb]{ .69,  .78,  .894} 49.2 & \cellcolor[rgb]{ .718,  .8,  .906} 44.8 & \cellcolor[rgb]{ .69,  .78,  .894} 49.4 & \cellcolor[rgb]{ .698,  .784,  .898} 48.4 & \cellcolor[rgb]{ .698,  .784,  .898} 48.4 & \cellcolor[rgb]{ .706,  .788,  .898} 47.2 & \cellcolor[rgb]{ .718,  .8,  .906} 45.0 & \cellcolor[rgb]{ .678,  .773,  .89} 51.0 & \cellcolor[rgb]{ .694,  .784,  .898} 48.8 & \cellcolor[rgb]{ .718,  .8,  .906} 45.2 & \cellcolor[rgb]{ .714,  .796,  .902} 45.8 & \cellcolor[rgb]{ .69,  .78,  .894} 49.6 & \cellcolor[rgb]{ .694,  .784,  .898} 48.6 \\
    WizardCoder-V1.0 & 15B   & \cellcolor[rgb]{ .843,  .89,  .949} 25.2 & \cellcolor[rgb]{ .812,  .867,  .937} 30.0 & \cellcolor[rgb]{ .808,  .867,  .937} 30.6 & \cellcolor[rgb]{ .792,  .855,  .929} 33.2 & \cellcolor[rgb]{ .835,  .882,  .945} 26.8 & \cellcolor[rgb]{ .792,  .851,  .929} 33.6 & \cellcolor[rgb]{ .812,  .867,  .937} 30.2 & \cellcolor[rgb]{ .812,  .867,  .937} 30.2 & \cellcolor[rgb]{ .808,  .863,  .933} 31.0 & \cellcolor[rgb]{ .796,  .855,  .929} 33.0 & \cellcolor[rgb]{ .788,  .851,  .929} 34.0 & \cellcolor[rgb]{ .804,  .859,  .933} 31.6 & \cellcolor[rgb]{ .816,  .871,  .937} 29.6 & \cellcolor[rgb]{ .796,  .855,  .929} 32.8 & \cellcolor[rgb]{ .804,  .863,  .933} 31.2 & \cellcolor[rgb]{ .804,  .863,  .933} 31.2 & \cellcolor[rgb]{ .816,  .867,  .937} 29.8 & \cellcolor[rgb]{ .788,  .847,  .929} 34.2 & \cellcolor[rgb]{ .788,  .851,  .929} 34.0 \\
    Starcoder & 15B   & \cellcolor[rgb]{ .875,  .91,  .957} 20.4 & \cellcolor[rgb]{ .804,  .859,  .933} 31.6 & \cellcolor[rgb]{ .8,  .859,  .933} 31.8 & \cellcolor[rgb]{ .808,  .863,  .933} 31.0 & \cellcolor[rgb]{ .886,  .922,  .961} 18.4 & \cellcolor[rgb]{ .792,  .851,  .929} 33.4 & \cellcolor[rgb]{ .8,  .859,  .933} 32.2 & \cellcolor[rgb]{ .8,  .859,  .933} 31.8 & \cellcolor[rgb]{ .816,  .867,  .937} 29.8 & \cellcolor[rgb]{ .796,  .855,  .929} 32.6 & \cellcolor[rgb]{ .796,  .855,  .929} 32.6 & \cellcolor[rgb]{ .812,  .867,  .937} 30.0 & \cellcolor[rgb]{ .82,  .871,  .937} 29.2 & \cellcolor[rgb]{ .792,  .851,  .929} 33.4 & \cellcolor[rgb]{ .796,  .855,  .929} 32.6 & \cellcolor[rgb]{ .812,  .867,  .937} 30.0 & \cellcolor[rgb]{ .812,  .867,  .937} 30.2 & \cellcolor[rgb]{ .796,  .855,  .929} 33.0 & \cellcolor[rgb]{ .796,  .855,  .929} 33.0 \\
    phi-3-instruct & 14B   & \cellcolor[rgb]{ .788,  .847,  .929} 34.2 & \cellcolor[rgb]{ .765,  .831,  .922} 37.6 & \cellcolor[rgb]{ .757,  .827,  .918} 39.0 & \cellcolor[rgb]{ .808,  .863,  .933} 31.0 & \cellcolor[rgb]{ .788,  .847,  .929} 34.2 & \cellcolor[rgb]{ .741,  .816,  .914} 41.6 & \cellcolor[rgb]{ .741,  .816,  .914} 41.2 & \cellcolor[rgb]{ .784,  .847,  .929} 34.4 & \cellcolor[rgb]{ .776,  .843,  .925} 35.8 & \cellcolor[rgb]{ .765,  .831,  .922} 37.8 & \cellcolor[rgb]{ .733,  .812,  .91} 42.4 & \cellcolor[rgb]{ .773,  .839,  .922} 36.6 & \cellcolor[rgb]{ .847,  .89,  .949} 24.6 & \cellcolor[rgb]{ .737,  .812,  .91} 42.2 & \cellcolor[rgb]{ .765,  .835,  .922} 37.4 & \cellcolor[rgb]{ .773,  .839,  .922} 36.2 & \cellcolor[rgb]{ .769,  .835,  .922} 37.2 & \cellcolor[rgb]{ .741,  .816,  .914} 41.4 & \cellcolor[rgb]{ .729,  .808,  .91} 43.0 \\
    Llama-3-Instruct & 8B    & \cellcolor[rgb]{ .8,  .859,  .933} 32.0 & \cellcolor[rgb]{ .808,  .863,  .933} 30.8 & \cellcolor[rgb]{ .804,  .863,  .933} 31.2 & \cellcolor[rgb]{ .804,  .863,  .933} 31.4 & \cellcolor[rgb]{ .843,  .89,  .949} 25.0 & \cellcolor[rgb]{ .78,  .843,  .925} 35.0 & \cellcolor[rgb]{ .804,  .863,  .933} 31.4 & \cellcolor[rgb]{ .788,  .851,  .929} 34.0 & \cellcolor[rgb]{ .816,  .871,  .937} 29.6 & \cellcolor[rgb]{ .831,  .882,  .945} 27.0 & \cellcolor[rgb]{ .792,  .851,  .929} 33.6 & \cellcolor[rgb]{ .831,  .878,  .945} 27.2 & \cellcolor[rgb]{ .827,  .878,  .941} 28.0 & \cellcolor[rgb]{ .8,  .859,  .933} 31.8 & \cellcolor[rgb]{ .784,  .847,  .929} 34.4 & \cellcolor[rgb]{ .788,  .851,  .929} 33.8 & \cellcolor[rgb]{ .8,  .859,  .933} 32.0 & \cellcolor[rgb]{ .773,  .839,  .922} 36.4 & \cellcolor[rgb]{ .788,  .851,  .929} 33.8 \\
    CodeQwen1.5-Chat & 7B    & \cellcolor[rgb]{ .765,  .831,  .922} 37.8 & \cellcolor[rgb]{ .749,  .824,  .914} 40.2 & \cellcolor[rgb]{ .749,  .824,  .914} 40.2 & \cellcolor[rgb]{ .745,  .82,  .914} 40.6 & \cellcolor[rgb]{ .78,  .843,  .925} 35.4 & \cellcolor[rgb]{ .725,  .808,  .906} 43.6 & \cellcolor[rgb]{ .733,  .812,  .91} 42.6 & \cellcolor[rgb]{ .749,  .82,  .914} 40.4 & \cellcolor[rgb]{ .753,  .824,  .918} 39.6 & \cellcolor[rgb]{ .729,  .808,  .91} 43.0 & \cellcolor[rgb]{ .741,  .816,  .914} 41.4 & \cellcolor[rgb]{ .761,  .831,  .918} 38.2 & \cellcolor[rgb]{ .757,  .827,  .918} 39.0 & \cellcolor[rgb]{ .722,  .8,  .906} 44.6 & \cellcolor[rgb]{ .737,  .812,  .91} 42.0 & \cellcolor[rgb]{ .78,  .843,  .925} 35.0 & \cellcolor[rgb]{ .761,  .831,  .918} 38.2 & \cellcolor[rgb]{ .725,  .804,  .906} 43.8 & \cellcolor[rgb]{ .737,  .812,  .91} 42.2 \\
    CodeLlama-Instruct & 7B    & \cellcolor[rgb]{ .8,  .859,  .933} 32.2 & \cellcolor[rgb]{ .776,  .843,  .925} 35.6 & \cellcolor[rgb]{ .784,  .847,  .929} 34.4 & \cellcolor[rgb]{ .78,  .843,  .925} 35.0 & \cellcolor[rgb]{ .847,  .894,  .949} 24.4 & \cellcolor[rgb]{ .761,  .831,  .918} 38.2 & \cellcolor[rgb]{ .78,  .843,  .925} 35.2 & \cellcolor[rgb]{ .8,  .859,  .933} 32.2 & \cellcolor[rgb]{ .788,  .847,  .929} 34.2 & \cellcolor[rgb]{ .776,  .839,  .925} 36.0 & \cellcolor[rgb]{ .78,  .843,  .925} 35.4 & \cellcolor[rgb]{ .8,  .859,  .933} 32.0 & \cellcolor[rgb]{ .816,  .871,  .937} 29.6 & \cellcolor[rgb]{ .769,  .835,  .922} 37.0 & \cellcolor[rgb]{ .765,  .835,  .922} 37.4 & \cellcolor[rgb]{ .796,  .855,  .929} 33.0 & \cellcolor[rgb]{ .796,  .855,  .929} 33.0 & \cellcolor[rgb]{ .784,  .847,  .925} 34.6 & \cellcolor[rgb]{ .757,  .827,  .918} 38.8 \\
    CodeLlama-hf & 7B    & \cellcolor[rgb]{ .796,  .855,  .929} 32.6 & \cellcolor[rgb]{ .784,  .847,  .929} 34.4 & \cellcolor[rgb]{ .788,  .851,  .929} 33.8 & \cellcolor[rgb]{ .792,  .851,  .929} 33.4 & \cellcolor[rgb]{ .824,  .875,  .941} 28.4 & \cellcolor[rgb]{ .761,  .831,  .922} 38.0 & \cellcolor[rgb]{ .78,  .843,  .925} 35.2 & \cellcolor[rgb]{ .784,  .847,  .929} 34.4 & \cellcolor[rgb]{ .78,  .843,  .925} 35.2 & \cellcolor[rgb]{ .761,  .831,  .922} 38.0 & \cellcolor[rgb]{ .784,  .847,  .929} 34.4 & \cellcolor[rgb]{ .796,  .855,  .929} 32.6 & \cellcolor[rgb]{ .808,  .863,  .933} 30.8 & \cellcolor[rgb]{ .784,  .847,  .925} 34.8 & \cellcolor[rgb]{ .769,  .835,  .922} 36.8 & \cellcolor[rgb]{ .792,  .851,  .929} 33.4 & \cellcolor[rgb]{ .808,  .863,  .933} 31.0 & \cellcolor[rgb]{ .78,  .843,  .925} 35.0 & \cellcolor[rgb]{ .761,  .831,  .918} 38.2 \\
    Deepseekcoder-instruct & 6.7B  & \cellcolor[rgb]{ .784,  .847,  .925} 34.8 & \cellcolor[rgb]{ .737,  .816,  .91} 41.8 & \cellcolor[rgb]{ .749,  .82,  .914} 40.4 & \cellcolor[rgb]{ .753,  .824,  .918} 39.4 & \cellcolor[rgb]{ .796,  .855,  .929} 32.8 & \cellcolor[rgb]{ .702,  .788,  .898} 47.6 & \cellcolor[rgb]{ .733,  .812,  .91} 42.6 & \cellcolor[rgb]{ .757,  .827,  .918} 38.8 & \cellcolor[rgb]{ .737,  .812,  .91} 42.0 & \cellcolor[rgb]{ .725,  .804,  .906} 43.8 & \cellcolor[rgb]{ .725,  .808,  .906} 43.6 & \cellcolor[rgb]{ .745,  .82,  .914} 40.8 & \cellcolor[rgb]{ .757,  .827,  .918} 39.2 & \cellcolor[rgb]{ .729,  .808,  .91} 43.2 & \cellcolor[rgb]{ .737,  .816,  .91} 41.8 & \cellcolor[rgb]{ .745,  .82,  .914} 40.6 & \cellcolor[rgb]{ .765,  .831,  .922} 37.8 & \cellcolor[rgb]{ .729,  .808,  .91} 43.2 & \cellcolor[rgb]{ .725,  .804,  .906} 44.0 \\
    Deepseekcoder-base & 6.7B  & \cellcolor[rgb]{ .741,  .816,  .914} 41.2 & \cellcolor[rgb]{ .71,  .796,  .902} 46.2 & \cellcolor[rgb]{ .729,  .808,  .91} 43.2 & \cellcolor[rgb]{ .733,  .812,  .91} 42.8 & \cellcolor[rgb]{ .733,  .812,  .91} 42.6 & \cellcolor[rgb]{ .718,  .8,  .906} 44.8 & \cellcolor[rgb]{ .714,  .796,  .902} 46.0 & \cellcolor[rgb]{ .745,  .82,  .914} 41.0 & \cellcolor[rgb]{ .749,  .82,  .914} 40.4 & \cellcolor[rgb]{ .737,  .816,  .91} 41.8 & \cellcolor[rgb]{ .718,  .8,  .906} 44.8 & \cellcolor[rgb]{ .733,  .812,  .91} 42.8 & \cellcolor[rgb]{ .729,  .808,  .91} 43.0 & \cellcolor[rgb]{ .733,  .812,  .91} 42.6 & \cellcolor[rgb]{ .737,  .812,  .91} 42.0 & \cellcolor[rgb]{ .729,  .808,  .91} 43.2 & \cellcolor[rgb]{ .745,  .82,  .914} 40.6 & \cellcolor[rgb]{ .702,  .788,  .898} 47.6 & \cellcolor[rgb]{ .718,  .8,  .902} 45.4 \\
    CodeGen-multi & 6B    & \cellcolor[rgb]{ .867,  .906,  .957} 21.4 & \cellcolor[rgb]{ .855,  .898,  .949} 23.6 & \cellcolor[rgb]{ .843,  .89,  .949} 25.0 & \cellcolor[rgb]{ .835,  .882,  .945} 26.4 & \cellcolor[rgb]{ .867,  .906,  .957} 21.6 & \cellcolor[rgb]{ .859,  .902,  .953} 22.8 & \cellcolor[rgb]{ .859,  .902,  .953} 22.8 & \cellcolor[rgb]{ .851,  .894,  .949} 23.8 & \cellcolor[rgb]{ .875,  .91,  .957} 20.4 & \cellcolor[rgb]{ .843,  .89,  .949} 25.2 & \cellcolor[rgb]{ .847,  .89,  .949} 24.8 & \cellcolor[rgb]{ .855,  .898,  .953} 23.4 & \cellcolor[rgb]{ .89,  .922,  .965} 17.8 & \cellcolor[rgb]{ .851,  .894,  .949} 24.0 & \cellcolor[rgb]{ .843,  .89,  .949} 25.2 & \cellcolor[rgb]{ .863,  .902,  .953} 22.0 & \cellcolor[rgb]{ .863,  .902,  .953} 22.2 & \cellcolor[rgb]{ .843,  .89,  .949} 25.0 & \cellcolor[rgb]{ .867,  .906,  .957} 21.4 \\
    phi-1\_5 & 1.3B  & \cellcolor[rgb]{ .902,  .929,  .969} 16.0 & \cellcolor[rgb]{ .839,  .886,  .945} 26.0 & \cellcolor[rgb]{ .847,  .89,  .949} 24.8 & \cellcolor[rgb]{ .859,  .902,  .953} 22.6 & \cellcolor[rgb]{ .902,  .933,  .969} 15.8 & \cellcolor[rgb]{ .859,  .898,  .953} 23.0 & \cellcolor[rgb]{ .855,  .898,  .949} 23.6 & \cellcolor[rgb]{ .871,  .906,  .957} 21.2 & \cellcolor[rgb]{ .863,  .902,  .953} 22.0 & \cellcolor[rgb]{ .863,  .902,  .953} 22.2 & \cellcolor[rgb]{ .839,  .886,  .945} 25.6 & \cellcolor[rgb]{ .867,  .906,  .953} 21.8 & \cellcolor[rgb]{ .898,  .925,  .965} 16.8 & \cellcolor[rgb]{ .878,  .914,  .961} 19.6 & \cellcolor[rgb]{ .863,  .902,  .953} 22.0 & \cellcolor[rgb]{ .867,  .906,  .957} 21.6 & \cellcolor[rgb]{ .89,  .922,  .965} 17.6 & \cellcolor[rgb]{ .839,  .886,  .945} 25.6 & \cellcolor[rgb]{ .843,  .89,  .949} 25.2 \\
    phi-1 & 1.3B  & \cellcolor[rgb]{ .965,  .976,  .988} 5.8 & \cellcolor[rgb]{ .945,  .961,  .984} 9.0 & \cellcolor[rgb]{ .922,  .945,  .973} 13.2 & \cellcolor[rgb]{ .91,  .937,  .969} 14.8 & \cellcolor[rgb]{ .973,  .98,  .992} 4.6 & \cellcolor[rgb]{ .871,  .91,  .957} 20.8 & \cellcolor[rgb]{ .882,  .918,  .961} 19.2 & \cellcolor[rgb]{ .902,  .933,  .969} 15.8 & \cellcolor[rgb]{ .906,  .933,  .969} 15.6 & \cellcolor[rgb]{ .886,  .918,  .961} 18.6 & \cellcolor[rgb]{ .863,  .902,  .953} 22.4 & \cellcolor[rgb]{ .89,  .922,  .965} 17.6 & \cellcolor[rgb]{ .937,  .957,  .98} 10.4 & \cellcolor[rgb]{ .89,  .922,  .965} 18.0 & \cellcolor[rgb]{ .898,  .929,  .965} 16.4 & \cellcolor[rgb]{ .933,  .953,  .976} 11.0 & \cellcolor[rgb]{ .898,  .929,  .965} 16.4 & \cellcolor[rgb]{ .882,  .918,  .961} 19.2 & \cellcolor[rgb]{ .882,  .918,  .961} 19.0 \\
    \midrule
    Average &       & \cellcolor[rgb]{ 1,  .933,  .733} 40.3 & \cellcolor[rgb]{ 1,  .922,  .675} 42.4 & \cellcolor[rgb]{ 1,  .922,  .678} 42.3 & \cellcolor[rgb]{ 1,  .922,  .671} 42.6 & \cellcolor[rgb]{ 1,  .949,  .8} 37.9 & \cellcolor[rgb]{ 1,  .902,  .6} 45.1 & \cellcolor[rgb]{ 1,  .918,  .659} 43.1 & \cellcolor[rgb]{ 1,  .925,  .698} 41.6 & \cellcolor[rgb]{ 1,  .929,  .706} 41.3 & \cellcolor[rgb]{ 1,  .914,  .635} 43.8 & \cellcolor[rgb]{ 1,  .91,  .624} 44.2 & \cellcolor[rgb]{ 1,  .933,  .733} 40.4 & \cellcolor[rgb]{ 1,  .949,  .792} 38.3 & \cellcolor[rgb]{ 1,  .914,  .643} 43.5 & \cellcolor[rgb]{ 1,  .914,  .651} 43.4 & \cellcolor[rgb]{ 1,  .925,  .702} 41.5 & \cellcolor[rgb]{ 1,  .937,  .741} 40.1 & \cellcolor[rgb]{ 1,  .906,  .616} 44.6 & \cellcolor[rgb]{ 1,  .906,  .616} 44.6 \\
    \bottomrule
    \end{tabular}%
    }
    \caption{The result of each LLM in \name. Each result is shaded with a background
color from blue to white based on the Pass@1. The bluer, the larger.  Deepseekcoder-V2 is a MOE LLM; the parameter activated during inference is 21B.}
  \label{tab:result}%
\end{table*}%


\noindent{\bf LLMs for evaluation} We selected 24 LLMs across 4 types for evaluation, including \textbf{\textit{general}} LLMs (GPT-3.5-Turbo, GPT-4o-mini, GPT-4o~\citep{brown2020language,achiam2023gpt}, Llama3~\citep{meta_llama_3}, Qwen2~\citep{yang2024qwen2}, phi-3-instruct~\citep{abdin2024phi}), \textbf{\textit{multilingual code}} LLMs (Deepseekcoder-V2~\citep{zhu2024deepseek}, Deepseekcoder-V1~\citep{guo2024deepseek}, CodeLlama~\citep{roziere2023code}, Starcoder~\citep{li2023starcoder}, Starcoder2~\citep{lozhkov2024starcoder}, CodeQwen1.5-Chat~\citep{bai2023qwen}), \textbf{\textit{instruction-tuned multilingual}} LLMs (Deepseekcoder-instruct-V1~\citep{guo2024deepseek}, WizardCoder~\citep{luo2023wizardcoder}, CodeLlama-Python, CodeLlama-Instruct~\citep{roziere2023code}), and \textbf{\textit{single or few-language code}} LLMs (CodeGen~\citep{nijkamp2022codegen}, phi-1~\citep{gunasekar2023textbooks}, phi-1.5~\citep{li2023textbooks}).

\noindent{\bf Evaluation task} We adopt the task settings from prior work~\citep{gu2024cruxeval}, dividing the tasks into \outputtask ~and \inputtask. For any PLs dataset, we provide the code along with the corresponding input or output in the test cases. \textbf{\textit{Input reasoning}} is predicting the input based on the output. \textbf{\textit{Output reasoning}} is predicting the output based on the input. 
{Each problem contains a test case, which follows the CRUXEVAL. The search space for test cases is exponential in nature, ensuring that LLMs do not know the answer to the problem in advance.}


\noindent{\bf Evaluation method.} We use pass@1 \citep{kulal2019spoc, chen2021evaluating} to evaluate both tasks. We set the temperature to 0 and employ greedy decoding for generation as prior work~\citep{javabench}. For closed-source LLMs, we use \textit{gpt-3.5-turbo}, \textit{gpt-4o-mini}, and \textit{gpt-4o} from OpenAI's API.

\subsection{Overall Result}
Table \ref{tab:result} shows Pass@1 evaluation results of various LLMs arranged in descending order of the LLMs' parameter size.
We can see that:
\textit{\textbf{1) \name~is challenging for all tested LLMs.}}
Even GPT-4o can only achieve Pass@1 around 70\%. Notably, the result of the open-source LLMs Deepseekcoder-V2 is better than GPT-4o-mini, with an average Pass@1 of 62.8\% on \inputtask ~and 65.0\% on \outputtask.
Furthermore, the average results show that across different types of PLs, LLMs exhibit similar capabilities in input and output reasoning, consistent with prior observations in Python~\citep{gu2024cruxeval}.
\textit{\textbf{2) Single-programming-language model (e.g., phi-1) and few-language models (e.g. CodeGen-muti) demonstrate unexpected generalization capabilities on unseen language tasks.}}
During evaluation, we introduced few-language LLMs (CodeGen-multi, trained on six PLs) and single-language LLMs (phi-1.5 and phi-1, trained only on Python).
Despite this, they achieved similar results across 19 PLs. Notably, phi-1, trained solely on Python, scored 11.8\% Pass@1 on Python input prediction and 23.6\% on Perl.
We will provide a more detailed analysis of this phenomenon in the Analysis section.

\section{Analysis}
\subsection{Key Factors for LLM Code Reasoning}
\label{sec:4.1}
\begin{figure}[t!]
\centering
\includegraphics[width=0.47\textwidth]{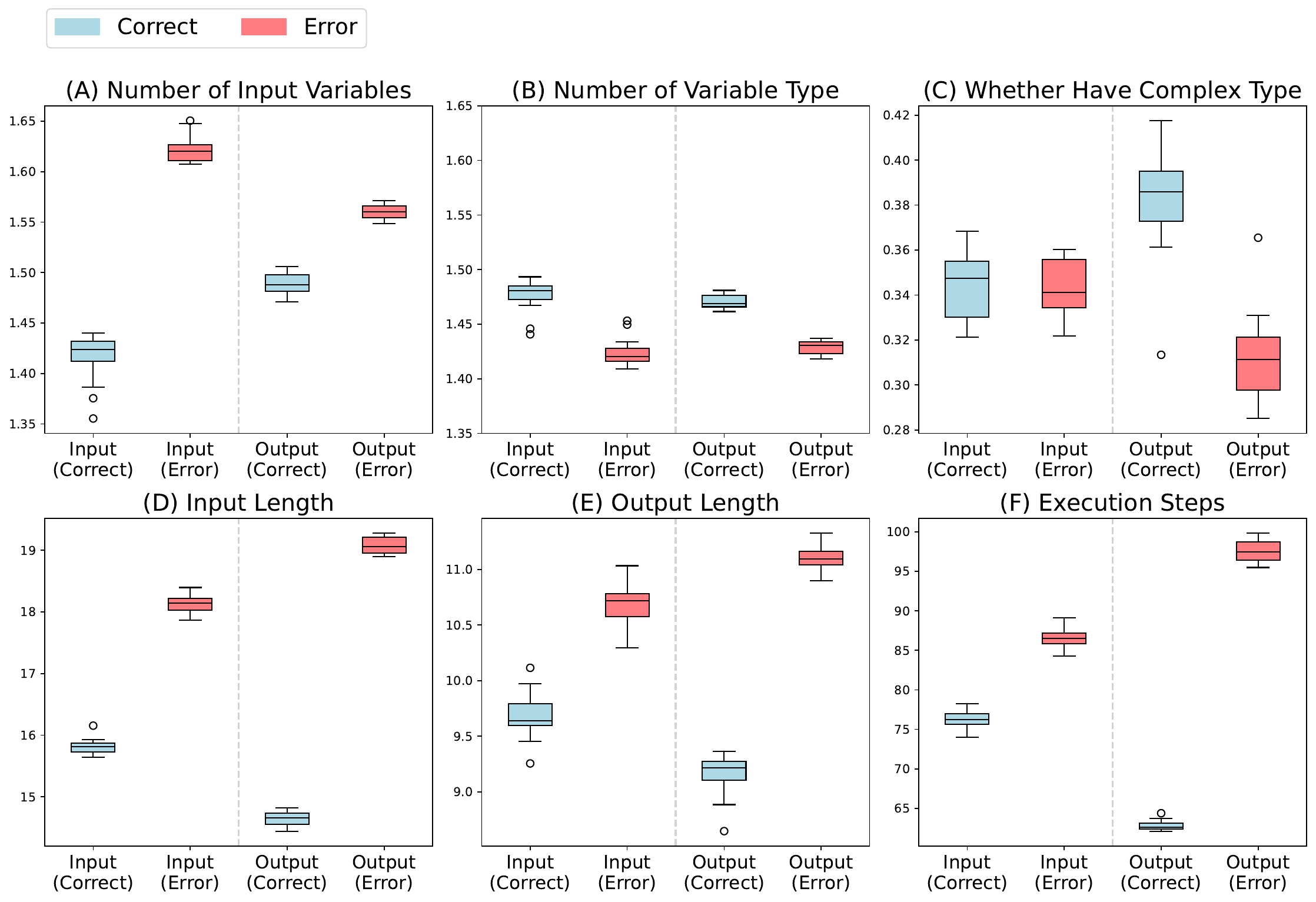}
\caption{Key factors for LLM code reasoning capability. 
}
\label{fig:difficulty}
\end{figure}

To get more insight into what factors in code affect LLMs' code reasoning ability, we explore six factors (\eg, average number of input variables, average input length) and statistics their correlation with correct/incorrect reasoning. 
The results are shown in Figure~\ref{fig:difficulty}. Each column of the box plot displays the distribution of 19 PLs. In particular, the columns ``{Num of Input Variable}'', ``{Num of Variable Type}'' are counted by averaging the number of types of input parameters in method signatures, ``{Input/Output Length}'' is the average string length of the input/output. For instance, in the test case where {\texttt{f('a', 123) == \{'a': 1\}}, the input length is 6, and the output length is 8}. ``{Whether Have Complex Type}'' checks whether there are \texttt{List}, \texttt{Dict}, \texttt{Tuple}, \texttt{Set} types in input and output. ``{Execution Steps}'' calculates the average execution steps in Python bytecode operations, following prior work~\citep{gu2024cruxeval}.

From Figure \ref{fig:difficulty} ( A - C ), we can see that \textbf{\textit{the number/types of input variables have little impact on the code reasoning}}, especially Sub-figure B, which shows that the reasoning capability is slightly better when more types of variables are involved. A more counter-intuitive observation is made from Figure \ref{fig:difficulty} ( D - E ). They indicate that \textbf{\textit{the reasoning capability is negatively correlated with the length of input/output strings}}.

Furthermore, regarding input reasoning capability, the more input variables, the more challenging it is for LLMs to reason about the correct inputs, thus the worse the input reasoning performance (Sub-figures A and D). 

\subsection{Cross-language Generalization Ability}

To investigate the cross-language generalization ability of LLMs, we investigate the reasoning ability of 
\textbf{\textit{phi-1}} and \textbf{\textit{phi-1.5}}, which are trained on English and Python only. To get a better understanding, we analyze the capability in terms of \textbf{\textit{syntax}} and \textbf{\textit{semantics}} in 9 PLs because they provide clear error messages to distinct syntactic/semantic errors.

\subsubsection{Syntactic Correctness.}{For a LLM trained on single PL, ensuring syntactic correctness is crucial for its generalization to other programming languages. For example, in a Racket language test case, the expected output might be "(list 5 8 1 3 0)", but the LLM incorrectly predicts "[5,8,1,3,0]", leading to a compilation error. }Figure~\ref{fig:synax} shows the number of syntactic-correct cases made by these two LLMs in both tasks. It is clear that \textbf{\textit{Python}} has the highest syntactic correctness in both tasks, followed by \textbf{\textit{Go}} and \textbf{\textit{TypeScript}}. On the contrary, C++, C\#, and Java witness the most syntactic errors for these LLMs. Interestingly, even though phi-1 and phi-1.5 have not trained on PLs other than Python, they can still achieve an average of 49.1\% and 72.0\% syntactically correctness rate in other PLs, respectively, compared with 97.0\% and 98.7\% achieved on Python. It indicates \textbf{\textit{the cross-language generalizability}} of LLMs.

\subsubsection{Semantic Correctness.} Beyond syntactic correctness, semantic correctness poses higher requirements, \ie, passing the tests. The results are shown in the last two rows in Table~\ref{tab:result}. In particular, phi-1.5 reaches 25.8\% input reasoning performance on Python, while on other PLs, an average of 19.0\% can also be reached. The observation further consolidates \textbf{\textit{the cross-language generalizability}} of LLMs.

\subsubsection{Cross-NL and Cross-PL Generalization.} 
\begin{figure}[t!]
\centering
\includegraphics[width=0.47\textwidth]{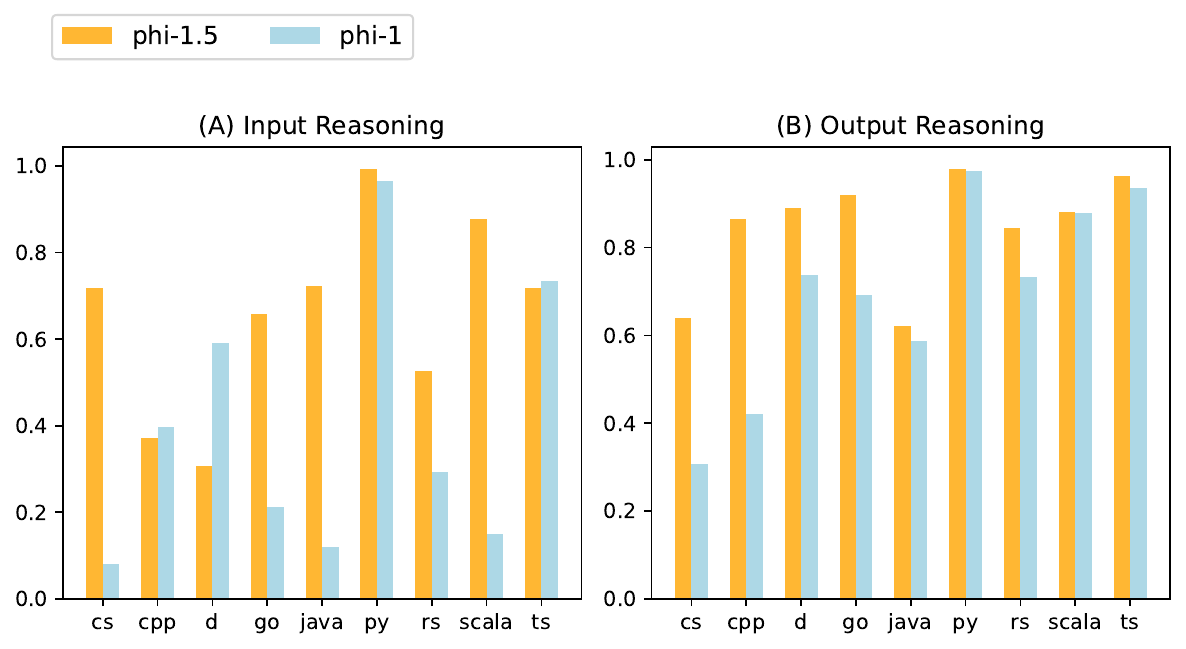}
\caption{The syntax accuracy of each LLM.}
\label{fig:synax}
\end{figure}

From Figure~\ref{fig:synax} and Table~\ref{tab:result}, there is a noticeable increase from phi-1 to phi-1.5 (an average of 10.7\% vs. 21.7\% on input reasoning, and 15.1\% vs. 21.7\% on output reasoning). According to the description~\citep{abdin2024phi}, phi-1.5 is further fine-tuned with more synthetic texts in \textbf{\textit{natural language}} (NL). Considering the dramatic improvement in code reasoning, it is highly likely that the improvement in NL reasoning positively impacts code reasoning. 

\subsection{Programming Language Correlation}

\begin{figure}[h!]
\centering
\includegraphics[width=1.0\linewidth]{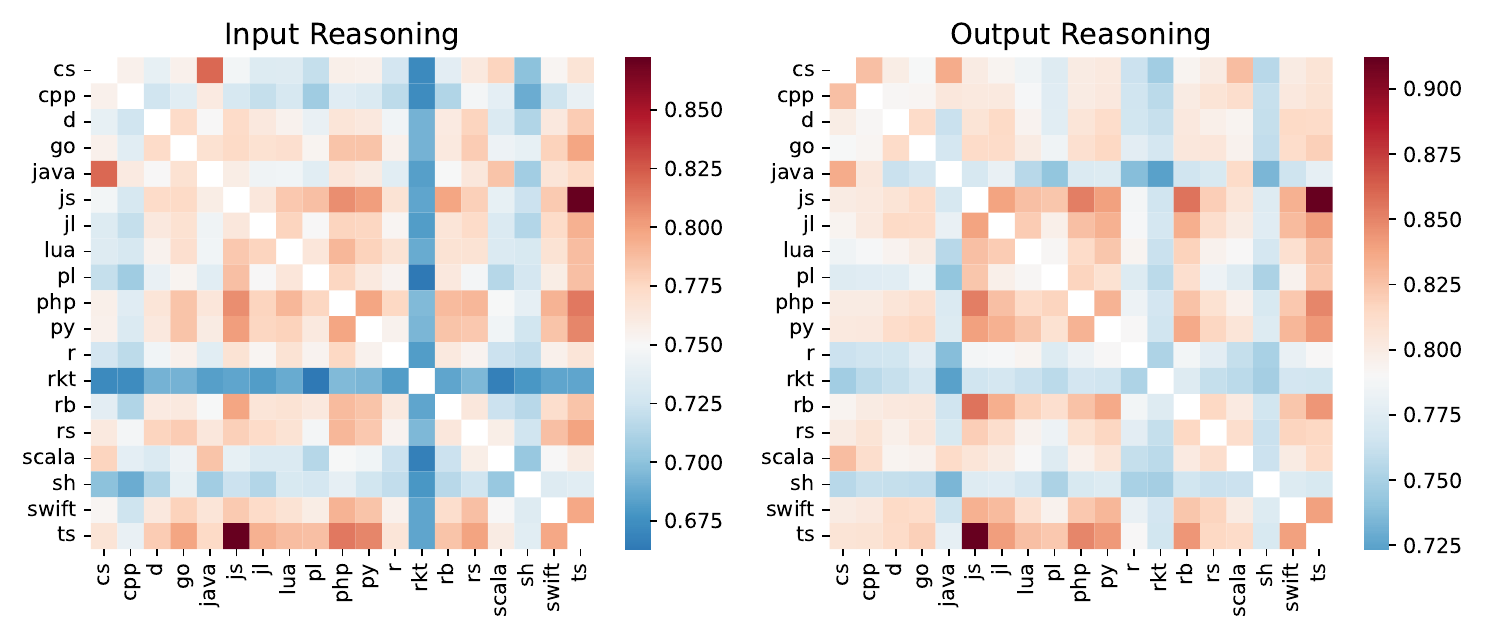}
\caption{The correlation between PL pairs.}
\label{fig:heatmap}
\end{figure}

To further investigate the correlations between these 19 PLs in \name, we calculate each PL pair's correlation (\ie, cosine similarity, ranging from -1 to 1), visualized in Figure~\ref{fig:heatmap}. In particular, for each PL, we flatten the results of LLMs as a feature vector and calculate the cosine similarities for each pair of PLs.  

Overall, Figure~\ref{fig:heatmap} shows that the correlation between PL pairs is generally similar, with an average of 0.7+ cosine similarities. Among all PL pairs, \textit{\textbf{JavaScript and TypeScript correlate the most strongly}} (0.87 and 0.91 on both tasks). 
It indicates that the \textbf{\textit{code reasoning capabilities on different PLs are highly correlated}}. Also, the correlation in output reasoning is slightly higher than in input reasoning, with an average of 0.79 vs. 0.75.

It is also noteworthy that {{Racket has the most minor correlation}} with all the other PLs. It may be because of its \textbf{\textit{distinct syntax}}. A case study can be found in Listing~\ref{lst:example3}.

\subsection{Case Study}

After identifying the phi-series-LLMs (\ie, phi-1, phi-1.5) exhibit cross-language generalization and the correlation across PLs, we further analyze the predictions of phi-1.5 to get a deeper understanding. We noticed that out of the 128 correct instances in Python by phi-1.5, 61.7\% (79/128) are also correct in PHP, while only 39.8\% (51/128) are correct in Racket. Therefore, we use one example in these three PLs to understand its rationale. 

\begin{listing}[h]%
\caption{Subject-164 (Python)}%
\label{lst:example1}%
\begin{lstlisting}[language=python]
def f(lst: List[int]) -> List[int]:    
    lst.sort()
    return lst[0:3]   
# input: [5, 8, 1, 3, 0]
# phi-1.5 answer: [0, 1, 3]
\end{lstlisting}
\end{listing}
\begin{listing}[h]%
\caption{Subject-164 (PHP)}%
\label{lst:example2}%
\begin{lstlisting}[language=php]
<?php
function f($lst) {
    sort($lst);
    return array_slice($lst, 0, 3);
}
// input: array(5, 8, 1, 3, 0)
// phi-1.5 answer: array(0, 1, 3)
\end{lstlisting}
\end{listing}
\begin{listing}[h!]%
\caption{Subject-164 (Racket)}%
\label{lst:example3}%
\begin{lstlisting}[]
(define (f lst)
  (define sorted-list (sort lst <))
  (take sorted-list 3))
;; input: (list 5 8 1 3 0)
;; phi-1.5 answer: (list 5 8 1)
\end{lstlisting}
\end{listing}

\subsubsection{Analysis on Subject 164.}
Listing~\ref{lst:example1}-\ref{lst:example3} demonstrates an instance where phi-1.5 generalizes Python (Listing~\ref{lst:example1})'s reasoning capabilities to other languages. From a grammar structure perspective, their respective function definitions, indentation formats, and the functions they invoke exhibit significant differences. However, overall, PHP and Python share a more similar structure, both utilizing \texttt{sort} for sorting and \texttt{return} for returning output values. Therefore, phi-1.5 is able to generalize its code reasoning abilities to PHP, but fails to comprehend the sorting command in Racket, leading to incorrect predictions.

Upon analyzing these 128 questions, we observe that excluding those where the output could be directly derived from the input, such as \texttt{assert f("zej","owc") == "zej"}, which accounted for approximately 40\% of the cases, there are still numerous examples demonstrating that phi-1.5 has developed a certain level of cross-language capabilities. From these examples, we can observe that the multilingual generalization capability of the model is positively correlated with the grammar structural similarity between languages. Even Racket, a language significantly different from others, maintains certain logical similarities in aspects such as function definitions, loops, and conditional branchs. This is a key reason why Phi-1.5 can achieve considerable effectiveness across multiple languages. { More case studies can be found in appendix \ref{sec:K}}.

\section{Related Work}

\noindent{\bf Multi-Task Code Benchmark.}
Recently, there has been an increasing number of tasks related to code that are used to evaluate the various capabilities of LLMs in the field of coding, including code generation~\citep{chen2021evaluating,austin2021program}, code repair~\citep{jimenez2023swe,tian2024debugbench}, and code description~\citep{chai2024mceval}. However, datasets that assess the reasoning abilities of code are relatively limited, and the currently proposed reasoning datasets are confined to the Python language~\citep{gu2024cruxeval,chen2024reasoningruntimebehaviorprogram}. In this work, we expand the Python language reasoning dataset CRUXEVAL~\citep{gu2024cruxeval} to encompass 19 PLs, thereby addressing the deficiency in reasoning datasets at the multilingual level.

\noindent{\bf Multi-Language Code Benchmark.}
Multilingual evaluation datasets are important for assessing the comprehensive coding capabilities of code LLMs. In the early stages, multilingual code datasets were mainly used for code translation tasks~\citep{elnaggar2021codetrans,ahmad2021avatar,roziere2020unsupervised,roziere2021leveraging,zhu2022multilingual,yan2023codetransocean,zhu2022xlcost}. These datasets often consist of problem solutions in different languages extracted from algorithm competition-related websites, thus suffering from data contamination issue. Bechmark like McEval~\citep{chai2024mceval}, which relies on human annotation, requires a high cost. In this work, we provide a process using LLMs for multilingual code translation, which can achieve a high accuracy and low cost in creating a multilingual dataset.

\section{Conclusion}
In this work, we provide a fully automated process for constructing a multilingual dataset based on a Python code language dataset. Through this process, we successfully transform the CRUXEval dataset into a multilingual dataset containing 19 PLs and test its effectiveness on 24 LLMs, demonstrating the validity of the dataset. Furthermore, we find that models trained on only a few languages exhibit the ability to transfer their prediction capabilities to other languages in input/output reasoning tasks, and this ability is influenced by the model's own reasoning capabilities.

\section{Limitations}
 There are Three limitations in this work. First, our benchmark relies on model-generated data. While this approach effectively reduces costs and avoids issues of data leakage, it may introduce biases in the dataset towards the LLMs that generated the data. However, we select different LLMs to construct evaluation data and conduct experiments on this data. The experimental results indicate that our dataset construction method does not affect the fairness of the dataset. Additionally, the model-based translation method cannot guarantee that all data will be perfectly translated into other PLs. However, through our method, based on 800 pieces of data, we generated 500 aligned pieces of data for 19 PLs. From our evaluation results, it can be seen that this amount of data is sufficient to effectively distinguish the reasoning capabilities of each LLM. {Finally, translated codes from Python cannot reflect several specific language features of other languages. Our intention is to achieve highly aligned multilingual data across 19 languages. However, preserving language-specific features make it challenging to create such aligned data, as the distinctive features of one language are inherently difficult to replicate in others. This represents a trade-off. In our future work, we plan to explore how to construct multilingual data while retaining more language-specific features.}
\bibliography{custom}

\appendix
\newpage
\section{Dataset Bias for Data-Generating LLM}
\label{sec:A}
%
The experimental results of existing papers have already demonstrated that the LLMs used for generate CRUXEVAL benchmark do not introduce unfairness \cite{gu2024cruxeval}. Therefore, our goal is to verify whether our pipeline can lead to unfairness on datasets of other programming languages (PLs). To achieve this, we reduce the number of iterations and use a single model for data generation, aiming to determine if different models can gain an advantage on the data they generated themselves.

We select Qwen2.5-Coder-32B-Instruct (Qwen 32B), deepseek-coder-33b-instruct (Dpk 33B), and deepseek-coder-6.7b-instruct (Dpk 7B) to generate data respectively, and reduce the maximum number of iterations to 5. We generated data for both commonly used PLs C++ and less commonly used PLs Racket. After taking the intersection, we obtain 566 aligned C++ data points and 206 aligned Racket data points. We then calculate the pass@1 for these three models, with the specific results shown in Table \ref{tab:bias}.
\begin{table}[h!]
  \centering
  \resizebox{0.48\textwidth}{!}{
    \begin{tabular}{c|ccc}
    \toprule
    \multicolumn{4}{c}{Input Reasoning (C++)} \\
    \midrule
    \diagbox{\small Data}{\small Evaluate}      & Qwen 32B & Dpk 33B & \multicolumn{1}{c}{Dpk 7B} \\
    \midrule
    Qwen 32B & \cellcolor[rgb]{ .388,  .745,  .482} 76.3 & \cellcolor[rgb]{ .804,  .914,  .843} 49.6 & \cellcolor[rgb]{ .98,  .984,  .992} 38.3 \\
    Dpk 33B & \cellcolor[rgb]{ .4,  .753,  .494} 75.6 & \cellcolor[rgb]{ .835,  .929,  .871} 47.5 & \cellcolor[rgb]{ .965,  .98,  .98} 39.2 \\
    Dpk 7B & \cellcolor[rgb]{ .388,  .745,  .482} 76.3 & \cellcolor[rgb]{ .827,  .925,  .863} 48 & \cellcolor[rgb]{ .988,  .988,  1} 37.6 \\
    \midrule
    \multicolumn{4}{c}{Output Reasoning (C++)} \\
    \midrule
     \diagbox{\small Data}{\small Evaluate}     & Qwen 32B & Dpk 33B & \multicolumn{1}{c}{Dpk 7B} \\
    \midrule
    Qwen 32B & \cellcolor[rgb]{ .443,  .769,  .529} 65.5 & \cellcolor[rgb]{ .741,  .89,  .788} 53.0 & \cellcolor[rgb]{ .988,  .988,  1} 42.6 \\
    Dpk 33B & \cellcolor[rgb]{ .388,  .745,  .482} 67.7 & \cellcolor[rgb]{ .784,  .906,  .824} 51.2 & \cellcolor[rgb]{ .949,  .973,  .969} 44.3 \\
    Dpk 7B & \cellcolor[rgb]{ .404,  .753,  .498} 67.1 & \cellcolor[rgb]{ .753,  .894,  .796} 52.6 & \cellcolor[rgb]{ .976,  .984,  .988} 43.2 \\
    \midrule
    \multicolumn{4}{c}{Input Reasoning (Racket)} \\
    \midrule
    \diagbox{\small Data}{\small Evaluate}      & Qwen 32B & Dpk 33B & \multicolumn{1}{c}{Dpk 7B} \\
    \midrule
    Qwen 32B & \cellcolor[rgb]{ .388,  .745,  .482} 82.0 & \cellcolor[rgb]{ .8,  .914,  .839} 51.9 & \cellcolor[rgb]{ .925,  .965,  .945} 42.7 \\
    Dpk 33B & \cellcolor[rgb]{ .455,  .773,  .541} 77.2 & \cellcolor[rgb]{ .839,  .929,  .871} 49.0 & \cellcolor[rgb]{ .988,  .988,  1} 37.9 \\
    Dpk 7B & \cellcolor[rgb]{ .435,  .765,  .525} 78.6 & \cellcolor[rgb]{ .933,  .965,  .953} 42.2 & \cellcolor[rgb]{ .945,  .973,  .961} 41.3 \\
    \midrule
    \multicolumn{4}{c}{Output Reasoning (Racket)} \\
    \midrule
     \diagbox{\small Data}{\small Evaluate}     & Qwen 32B & Dpk 33B & \multicolumn{1}{c}{Dpk 7B} \\
    \midrule
    Qwen 32B & \cellcolor[rgb]{ .4,  .749,  .49} 78.6 & \cellcolor[rgb]{ .745,  .89,  .792} 58.7 & \cellcolor[rgb]{ .941,  .969,  .957} 47.6 \\
    Dpk 33B & \cellcolor[rgb]{ .439,  .769,  .529} 76.2 & \cellcolor[rgb]{ .761,  .898,  .804} 57.8 & \cellcolor[rgb]{ .965,  .98,  .98} 46.1 \\
    Dpk 7B & \cellcolor[rgb]{ .388,  .745,  .482} 79.1 & \cellcolor[rgb]{ .78,  .906,  .82} 56.8 & \cellcolor[rgb]{ .988,  .988,  1} 44.7 \\
    \bottomrule
    \end{tabular}%
    }
    \caption{Impact of data generating LLMs}
  \label{tab:bias}%
\end{table}%

It can be observed that the relative ordering of the three models does not change. Moreover, the model used for data generation does not significantly impact the final evaluation results. For instance, in the Input Reasoning task for C++, Qwen 32B's pass@1 consistently remain around 76.

We analyze the possible reason: \textbf{During both the data generation phase and the evaluation phase, the tasks performed by the LLMs are different}. We use the model to generate code in different languages (i.e., code translation), while we evaluate the model's ability to infer outputs/inputs based on given inputs/outputs in code (i.e., code reasoning). It is important to note that the test cases used during the code reasoning phase are not visible to the model during the code translation phase, ensuring a fair evaluation process. Therefore, the model is performing two distinct tasks (i.e., code translation and code reasoning), and there is no unfairness involved.

\section{The Improvements of Pipline}
\label{sec:B}
\subsection{enhance of the pipline}
(1) For C\# we enhance the check function to include the ability to judge if the \texttt{List} and \texttt{Dict} types are equal.
(2) For Julia we add the data type to empty dict, for example change input \texttt{Dict()} to \texttt{Dict\{String, String\}()}. Otherwise, the function can not accept the input
(3) For JavaScript we change \texttt{String} which Contains special characters, such as\texttt{"example\textbackslash n"} to \texttt{\textasciigrave example\textbackslash n \textasciigrave}.
\subsection{transform complex types}(1) For inputs and outputs that include functions, such as \texttt{"bfrerat".split("-")}, we will replace the input or output with the result after the function execution. (2) When the input is of \texttt{Callable} type, such as \texttt{lambda x: x.reverse()}, we will remove the parameter and incorporate the \texttt{Callable} type into the main function internally. (3) For complex variables that contain multiple types, if we can convert them into a simpler type without altering the function's functionality, we will preserve such functions. For instance, as illustrated in Listing \ref{lst:example_complex_type}, consider a dictionary \texttt{d: Dict[str, Union[int, str]]}. If converting all its values to the \texttt{str} type does not alter the function's behavior, we will retain it; otherwise, we will discard them.
\begin{listing}[h]%
\caption{An example with complex type}%
\label{lst:example_complex_type}%
\begin{lstlisting}[language=python]
from typing import Dict, Union, Tuple

def f(d: Dict[str, Union[int, str]]) -> Tuple[bool, bool]:
    r = {
        "c": d.copy(),
        "d": d.copy()
    }
    return (r["c"] is r["d"], r["c"] == r["d"])

def check(candidate):
    assert candidate({"i": 1, "love": "parakeets"}) == (False, True)

def test_check():
    check(f)
\end{lstlisting}
\end{listing}
\section{The Difficulties for Translating}
\label{sec:C}

(1) For indexing functions, the starting position is not 0 but 1, with Julia being a typical language that exhibits this behavior. {An example is shown in listing \ref{lst:Julia Error Case}.}
(2) For the conversion between Python's \textit{str} type and its own \textit{char} and \textit{string} types, D language is a typical case where this issue arises. {An example is shown in listing \ref{lst:example1 for GPT-4o}.}
(3) For the transformation and comparison of dictionary types. {For example, C\# need to first sort two Dictionary before compare them.}

\begin{listing}[h!]%
\caption{Julia Error Case}%
\label{lst:Julia Error Case}%
\begin{lstlisting}[language=Python]
function f(text::String, s::Int64, e::Int64)::Int64
    sublist = text[s+1:e]  # Julia uses 1-based indexing, so adjust the start index
    if isempty(sublist)
        return -1
    end
    min_char = minimum(sublist)
    return findfirst(==(min_char), sublist) - 1  # Adjust for 0-based index in the result
end
\end{lstlisting}
\end{listing}

\section{Details of Benchmark Construction}
\label{sec:D}
\subsection{experimental setup of each step}
In Generation step. Given the higher usage cost of closed-source LLMs, we set $N$ to $5$, $k$ to $5$, $\delta A$ to $0$, and the temperature to 0.2 for GPT3.5-Turbo. For DeepseekCoder-33B-Instruct We set $N$ to $50$, $k$ to $5$, $\delta A$ to $0$, and the temperature to 0.8. In Repair step. We set temperature to 0. For GPT-4o, we generate once, and correct errors three times.
\subsection{prompt of each step}
\begin{figure}[h!]
\centering
\includegraphics[width=0.45\textwidth]{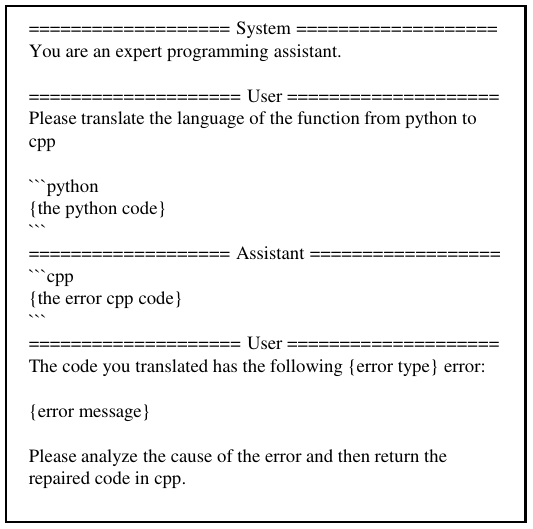}
\caption{Prompt of repair}
\label{fig:repair}
\end{figure}

\begin{figure}[h!]
\centering
\includegraphics[width=0.45\textwidth]{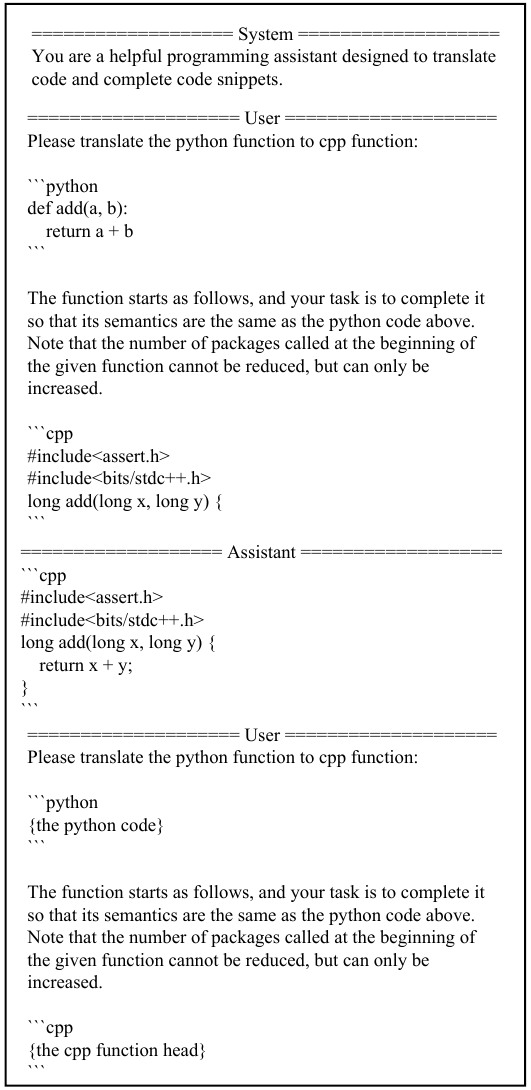}
\caption{Prompt of generation}
\label{fig:generation}
\end{figure}

In Figures \ref{fig:generation} and \ref{fig:repair}, we include the prompts we use for our benchmark construction. We use a few-shot prompt for all models. For Generation step of each model, the prompt is show in Figure \ref{fig:generation}. The example in this Figure is used for GPT3.5-turbo. After the Generation and Repair of GPT3.5-turbo, we choose three examples from the correct generated problems which include \texttt{str}, \texttt{List}, \texttt{Dict} type respectively. We use these examples for Deepseekcoder-instruct-33b Generation. For GPT-4o, based on the summarized difficulties in translation, we provide three examples, as shown in Listing \ref{lst:example1 for GPT-4o}, \ref{lst:example2 for GPT-4o}, and \ref{lst:example3 for GPT-4o}. we present these examples in three distinct languages. For each specific language translation, we employ the corresponding language version of the three examples.

As illustrated in Figure \ref{fig:repair}, the prompt for Repair is shown, which depicts a single round of error correction. The compiler's returned error messages are provided to the model for correction. For multiple rounds of error correction, subsequent error messages are appended to the dialogue after the model encounters errors again.
\begin{listing}[h!]%
\caption{example1 for GPT-4o (D)}%
\label{lst:example1 for GPT-4o}%
\begin{lstlisting}[language=python]
import std.math;
import std.typecons;
import std.conv;
import std.algorithm;
import std.array;
import std.string;

string f(string x, string y) {
    char[] yMutable = y.dup;
    yMutable.reverse();
    string tmp = yMutable.map!(c => c == "9" ? "0" : "9").array.map!(c => c.to!string).array.join("");
    if (x.isNumeric && tmp.isNumeric) {
        return x ~ tmp;
    } else {
        return x;
    }
}
unittest
{
    alias candidate = f; 
    assert(candidate("", "sdasdnakjsda80") == "");
}
void main(){}
\end{lstlisting}
\end{listing}


\begin{listing}[h!]%
\caption{example2 for GPT-4o (Swift)}%
\label{lst:example2 for GPT-4o}%
\begin{lstlisting}[language=python]
import Foundation

func f(strand: String, zmnc: String) -> Int {    
    var strand = strand
    var poz = strand.range(of: zmnc)
    while poz != nil {
        strand.removeSubrange(poz!)
        poz = strand.range(of: zmnc)
    }
    let lastIndex = strand.range(of: zmnc, options: [], range: nil, locale: nil)?.lowerBound.utf16Offset(in: strand)
    return lastIndex ?? -1

func ==(left: [(Int, Int)], right: [(Int, Int)]) -> Bool {
    if left.count != right.count {
        return false
    }
    for (l, r) in zip(left, right) {
        if l != r {
            return false
        }
    }
    return true
}
            
assert(f(strand: "", zmnc: "abc") == -1)
\end{lstlisting}
\end{listing}
\begin{listing}[h!]%
\caption{example3 for GPT-4o (Python)}%
\label{lst:example3 for GPT-4o}%
\begin{lstlisting}[language=python]
from typing import Dict,Tuple

def f(d: Dict[str, int]) -> Tuple[int,int]:
    if "x" in d:
        x = d["x"]
    if "y" in d:
        y = d["y"]
    return x,y

def check(candidate):
    assert candidate({"x": 5, "y": 12}) == (5, 12)
    
def test_check():
    check(f)
\end{lstlisting}
\end{listing}

\section{Programming Language Correlation in Translation}

We observed that after a four-step translation process, the intersection of the 18 programming languages contained only 333 entries. This indicates that each programming language has its unique subset of correctly translated parts. Therefore, we constructed a Venn diagram to study the correlation between the sets of correct translations among different languages.

\begin{figure}[h!]
\centering
\includegraphics[width=0.47\textwidth]{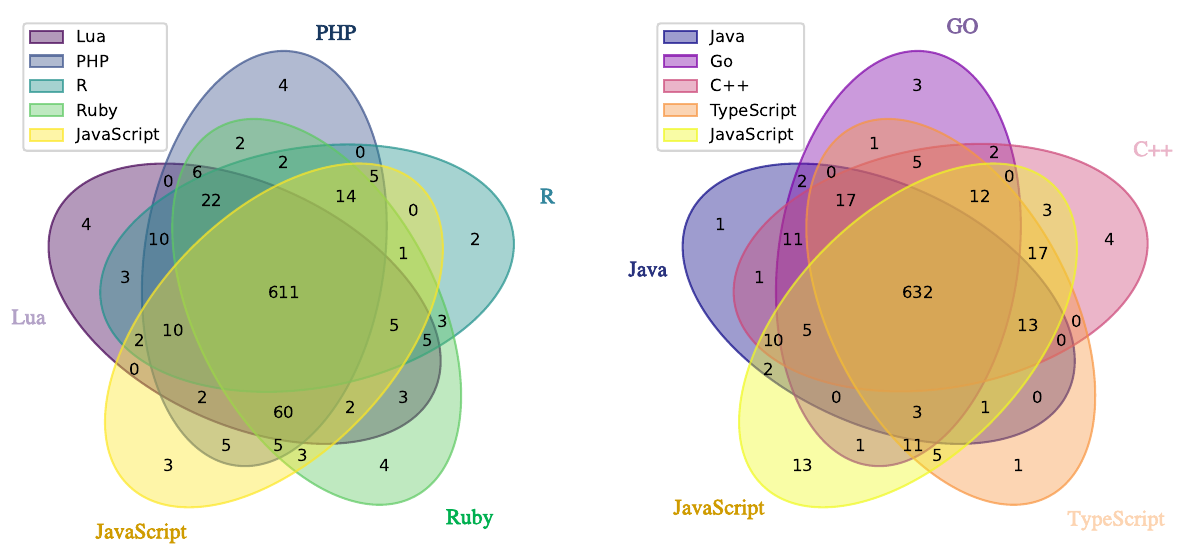}
\caption{The communitie of each langugage in code translation.}
\label{fig:translation connection}
\end{figure}

Specifically, the results are shown in Figure  \ref{fig:translation connection}. Due to the limitation of the number of sets that a Venn diagram can clearly represent, we explored by two methods, each selecting five representative languages. First, we chose the five languages with the largest union of results from the 18 languages to construct the first Venn diagram, which is the left half of Figure  \ref{fig:translation connection}. Subsequently, we selected the top five most widely used languages, excluding Python, based on data from GitHut 2.0, to construct the second Venn diagram, which is the right half of Figure  \ref{fig:translation connection}.

From Figure \ref{fig:translation connection}, we can observe that mainstream programming languages often have more similar syntax structures, and the model's generation capability is stronger for these languages. Therefore, the intersection of the generation results for these five languages is relatively large, with 632 entries, while the union is relatively small, totaling 776 entries. Lua, PHP, R, Ruby, and JavaScript are among the languages with the broadest correct translation entries across all programming languages, with their union totaling 798 entries.

\begin{figure}[ht]
\centering
\includegraphics[width=0.47\textwidth]{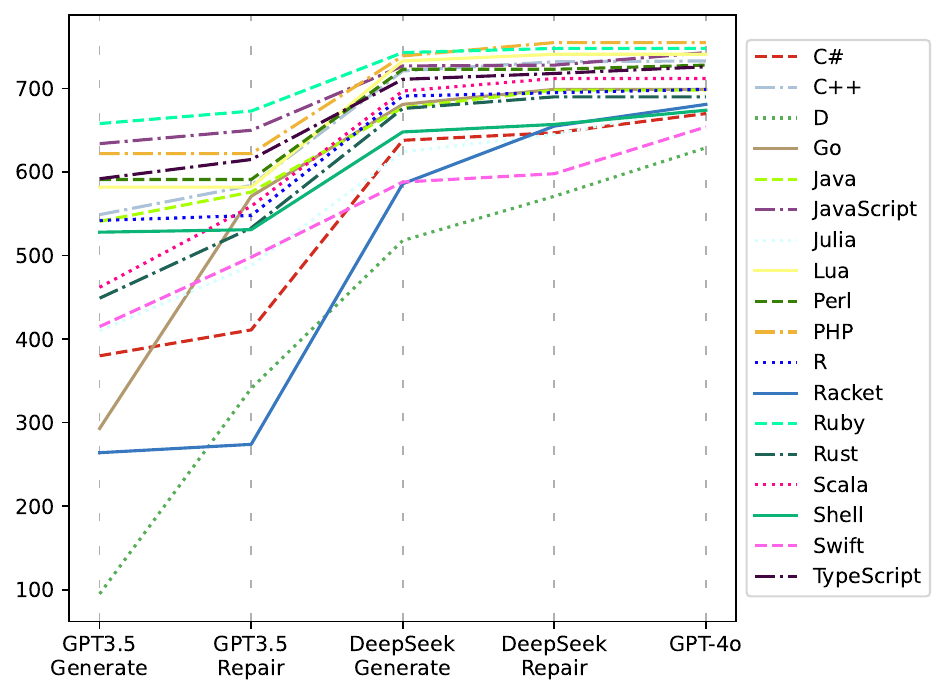}
\caption{The result of each step in translation}
\label{fig:iter_translate}
\end{figure}
\section{The Result of Each Step in Translation}

Figure \ref{fig:iter_translate} shows the improvement brought by each step for every language during the construction of the benchmark. It can be observed that each step leads to an overall enhancement in translation performance. For mainstream languages such as C++ and Java, the number of correctly translated items can exceed 600 after one or two steps. For lower-frequency languages like D and Racket, the effect is gradually improved, eventually resulting in all languages having more than 600 correct translations.
\section{GPU Usage and Total Cost of Translation}
The total cost of GPT3.5-Turbo and GPT-4o is aboat \$60 US dollars. For Deepseekcoder-Instruct-33b, we use 1 NVIDIA A100-80GB GPU and the generation and repair takes about 72 hours.

\section{Prompt of Input/Output Reasoning}
\begin{figure}[h!]
\centering
\includegraphics[width=0.47\textwidth]{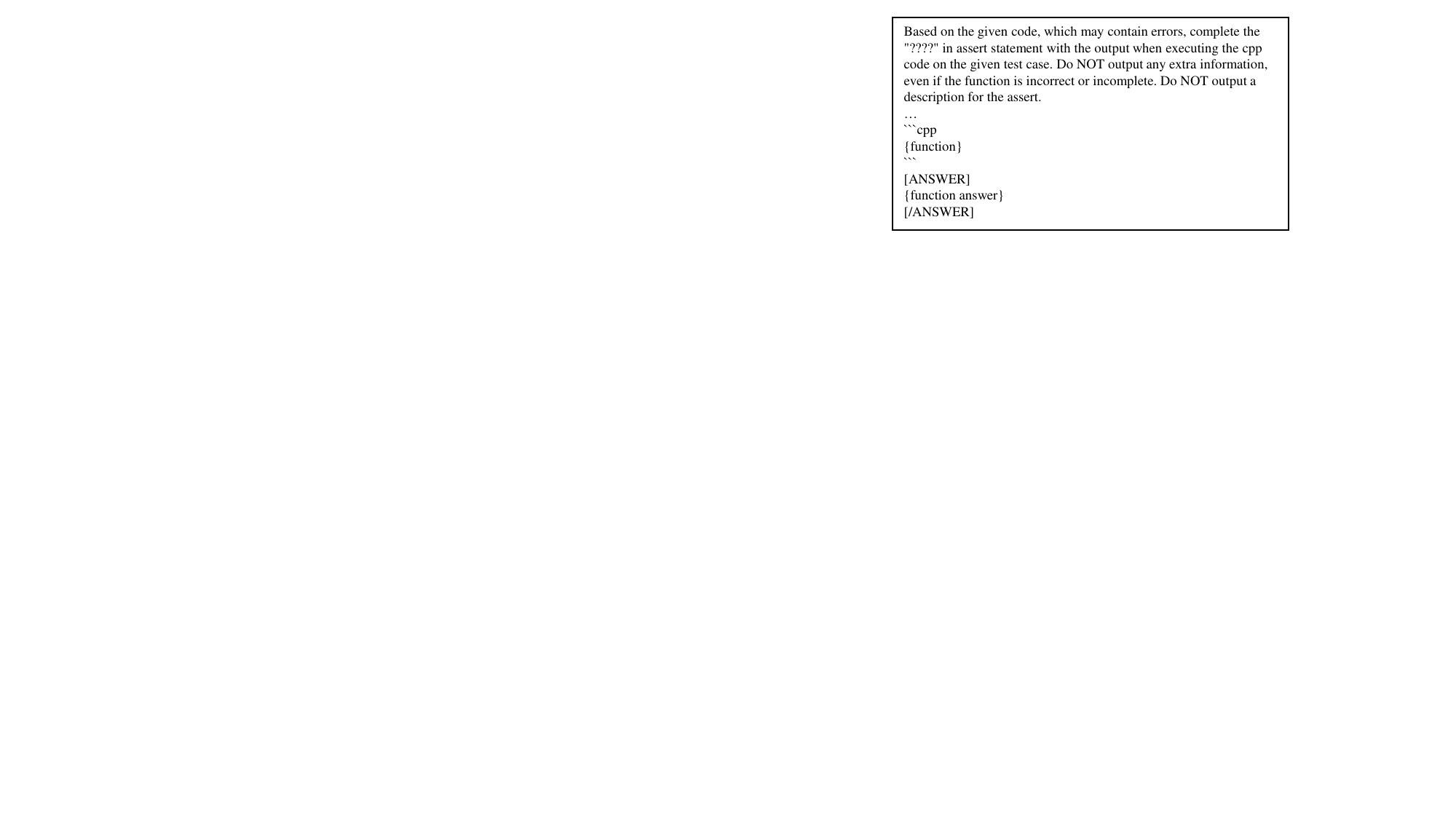}
\caption{Prompt of output reasoning (non-GPT)}
\label{fig:infer_output}
\end{figure}

\begin{figure}[h!]
\centering
\includegraphics[width=0.47\textwidth]{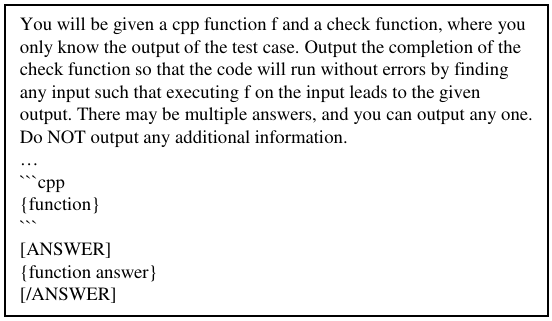}
\caption{Prompt of input reasoning (GPT)}
\label{fig:infer_input_gpt}
\end{figure}

\begin{figure}[h!]
\centering
\includegraphics[width=0.46\textwidth]
{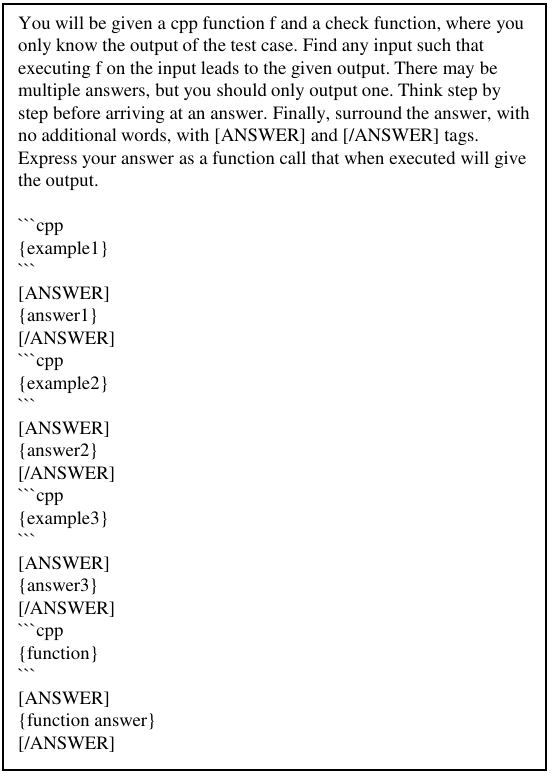}
\caption{Prompt of input reasoning (non-GPT)}
\label{fig:infer_input}
\end{figure}

\begin{figure}[h!]
\centering
\includegraphics[width=0.46\textwidth]{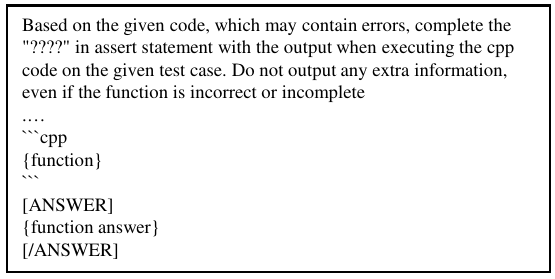}
\caption{Prompt of output reasoning (GPT)}
\label{fig:infer_output_gpt}
\end{figure}

In Figures \ref{fig:infer_input}, \ref{fig:infer_output}, \ref{fig:infer_input_gpt}, \ref{fig:infer_output_gpt}, we include the prompts we use for our benchmark construction. We use a few-shot prompt for all models. The examples of few-shot is shown in Listings \ref{lst:Input/Output Reasoning example1}, \ref{lst:Input/Output Reasoning example2}, \ref{lst:Input/Output Reasoning example3}. All the prompts and examples are demonstrated in the C++ language.
\begin{listing}[h!]%
\caption{Input/Output Reasoning example1 (C++)}%
\label{lst:Input/Output Reasoning example1}%
\begin{lstlisting}[language=C++]
#include<assert.h>
#include<bits/stdc++.h>
long f(std::vector<std::string> my_list) {
    long count = 0;
    for (std::string i : my_list) {
        if (i.size() % 2 == 0) {
            count += 1;
        }
    }
    return count;
}
int main() {
    auto candidate = f;
    assert(candidate((std::vector<std::string>({(std::string)"mq", (std::string)"px", (std::string)"zy"}))) == (3));
}
\end{lstlisting}
\end{listing}

\begin{listing}[h!]%
\caption{Input/Output Reasoning example2 (C++)}%
\label{lst:Input/Output Reasoning example2}%
\begin{lstlisting}[language=C++]
#include<assert.h>
#include<bits/stdc++.h>
std::string f(std::string s1, std::string s2) {
    return s1 + s2;
}
int main() {
    auto candidate = f;
    assert(candidate(("ba"), ("nana")) == ("banana"));
}
\end{lstlisting}
\end{listing}
\begin{listing}[h!]%
\caption{Input/Output Reasoning example3 (C++)}%
\label{lst:Input/Output Reasoning example3}%
\begin{lstlisting}[language=C++]
#include<assert.h>
#include<bits/stdc++.h>
std::tuple<long, long> f(std::map<std::string,long> d) {
    long x = 0, y = 0;
    if(d.find("x") != d.end()){
        x = d["x"];
    }
    if(d.find("y") != d.end()){
        y = d["y"];
    }
    return std::make_tuple(x, y);
}
int main() {
    auto candidate = f;
    assert(candidate((std::map<std::string,long>({{"x", 5}, {"y", 12}}))) == (std::make_tuple(5, 12)));
} 
\end{lstlisting}
\end{listing}
{
\section{The Statistics of Benchmark}
\label{sec:I}
We present specific statistics for our benchmark across various PLs. As shown in Table \ref{tab:statistics}, the average number of lines and average length for each PLs is provided. Each entry corresponds to one test case, and the input-output reasoning is evaluated by predicting the outcome of that test case.
}
\begin{table}[htbp]
  \centering
    \resizebox{0.48\textwidth}{!}{
    \begin{tabular}{ccccc}
    \toprule
    \textbf{language} & \textbf{Rust} & \textbf{Racket} & \textbf{Lua} & \textbf{Java} \\
    \midrule
    \textbf{avg line} & 14.1  & 17.6  & 17.3  & 22 \\
    \textbf{avg length} & 380.4 & 401.1 & 413.3 & 641.6 \\
    \midrule
    \textbf{Scala} & \textbf{PHP} & \textbf{Perl} & \textbf{GO} & \textbf{D} \\
    16.1  & 17.3  & 21.3  & 35.8  & 21.5 \\
    445.9 & 407.8 & 409.3 & 755.2 & 396.6 \\
    \midrule
    \textbf{C++} & \textbf{C\#} & \textbf{TS} & \textbf{Python} & \textbf{R} \\
    \midrule
    16.4  & 22.9  & 17.1  & 14.3  & 11.3 \\
    481.1 & 636.3 & 405.1 & 303.4 & 347.2 \\
    \midrule
    \textbf{Julia} & \textbf{Ruby} & \textbf{JS} & \textbf{Swift} & \textbf{Shell} \\
    15.2  & 14.5  & 16    & 24.5  & 23.2 \\
    299.3 & 321.9 & 344.6 & 551.4 & 370.4 \\
    \bottomrule
    \end{tabular}%
    }
    \caption{The average number of lines and average length of our benchmark}
  \label{tab:statistics}%
\end{table}%
{
\section{Code Reasoning Task}
\label{sec:J}
\subsection{The connection between code reasoning and input/output prediction}
Input/output prediction is a good carrier for demonstrating code reasoning capabilities. First, for code tasks such as code generation, code repair, and code translation, they all fundamentally depend on understanding the code execution process. The input/output reasoning ability of a LLM is excellent representation of this understanding. What's more, starting from the perspective of LLMs input/output reasoning is a task that is easy to judge for correctness and has a certain level of difficulty, requiring rigorous logical reasoning. This task is of great reference value for evaluating the reasoning capabilities of language models.
}
{
\subsection{The applicability of code reasoning tasks}
In software development, understanding and predicting the input and output of code is crucial for detecting potential security vulnerabilities and ensuring the reliability of the code. For example, input validation and output filtering are key steps in preventing security vulnerabilities such as SQL injection and cross-site scripting attacks. A LLM capable of precise input and output reasoning can provide strong support in this regard.
}
{
\section{Additional Case Study}
\label{sec:K}
\subsection{error cases of GPT-4o}
As can be seen in Table \ref{tab:result}, even GPT-4o fails to successfully solve approximately 30\% of the problems. We analyze this portion of cases, specifically selecting those where less than two PLs out of 19 passed the test cases. We find that a significant number of these cases involve the processing of long strings. In input reasoning tasks, this accounted for 85.2\% (46/54) of the cases. In Output reasoning tasks, it accounted for 68.4\% (39/57).
}

{
Listings \ref{lst:Error Case of GPT-4o (Input Reasoning, C++)} and \ref{lst:Error Case of GPT-4o (Output Reasoning, Python)} respectively show the error cases of GPT-4o in input / output reasoning tasks. In Listing \ref{lst:Error Case of GPT-4o (Input Reasoning, C++)}, the LLM fail to comprehend the code's intention of converting the case of every alternate character. In Listing \ref{lst:Error Case of GPT-4o (Output Reasoning, Python)}, the variable \textit{space\_symbol} consists of two parentheses, but the LLMs mistakenly interpreted it as one parenthesis, leading to an error. This aligns with the conclusion in Section \ref{sec:4.1} that the longer the input or output, the more prone the LLMs are to making mistakes.
}

{
\subsection{comparative evaluation of LLMs' reasoning ability}
To better understand the reasoning capabilities of different LLMs, we select three LLMs at varying levels: GPT-4o, DeepseekCoder-6.7b-base, and phi-1.5. We compare them pairwise to assess the reasoning ability of each LLM. We define that if a model correctly solves a problem in more than 17 PLs, it is considered to have mastered that problem. Conversely, if it solves the problem in fewer than 3 languages, it is deemed not to have mastered it.
}

{
By comparing the problems that GPT-4o master but DeepseekCoder-6.7b-base do not, we find that these problems mostly require multi-step operations, but the overall difficulty does not exceed the problem in Listing \ref{lst:Error Case of GPT-4o (Input Reasoning, C++)}. As shown in Listings \ref{lst:Correct Case of GPT-4o (Output Reasoning, Python)} and \ref{lst:Error Case of DeepseekCoder-6.7b-base (Output Reasoning, Python)}, one such problem involves successfully reversing '\ \ \ OOP\ \ \ ' and removing extra spaces. Although DeepseekCoder-6.7b-base also execute these two steps, it fail to completely remove the spaces on both sides.
}

{
Next, by comparing the problems that DeepseekCoder-6.7b-base master but phi-1.5 do not, we find that these issues often arise because phi-1.5 fails to correctly write variables corresponding to programming languages other than Python, accounting for 90\% (36/40) of the cases. As illustrated in Listings \ref{lst:Correct Case of DeepseekCoder-6.7b-base (Input Reasoning, C++)} and \ref{lst:Error Case of Phi-1_5 (Input Reasoning, C++)}, phi-1.5 does not understand the meaning of the long-type variable and still outputs it as a string, leading to syntax errors.
}

{
From the above analysis, it can be seen that GPT-4o is capable of handling relatively complex tasks, DeepseekCoder-6.7b-base performs well in simple tasks but may make minor errors in complex tasks. phi-1.5, although demonstrating a certain level of Cross-PL Generalization, still performs poorly in non-Python languages. Particularly in understanding variable types and syntax
}

\begin{listing}[h!]%
\caption{Error Case of GPT-4o (Input Reasoning, C++)}%
\label{lst:Error Case of GPT-4o (Input Reasoning, C++)}%
\begin{lstlisting}[language=C++]
#include<assert.h>
#include<bits/stdc++.h>
std::string f(std::string line) {
    int count = 0;
    std::string a;
    for (int i = 0; i < line.length(); i++) {
        count += 1;
        if (count%2==0) {
            a.push_back(tolower(line[i]) == line[i] ? toupper(line[i]) : tolower(line[i]));
        }
        else {
            a.push_back(line[i]);
        }
    }
    return a;
}
int main() {
    auto candidate = f;
    assert(candidate(("987yHnShAsHd 93275YrGsGbGsShFbSfB")) == ("987YhnShAShD 93275yRgsgBgssHfBsFB"));
}
\end{lstlisting}
\end{listing}

\begin{listing}[h!]%
\caption{Error Case of GPT-4o (Output Reasoning, Python)}%
\label{lst:Error Case of GPT-4o (Output Reasoning, Python)}%
\begin{lstlisting}[language=Python]
def f(text: str, space_symbol: str, size: int) -> str:    
    spaces = ''.join(space_symbol for i in range(size-len(text)))
    return text + spaces

def check(candidate):
    assert candidate('w', '))', 7) == 'w))))))'

def test_check():
    check(f)

test_check()
\end{lstlisting}
\end{listing}

\begin{listing}[h!]%
\caption{Correct Case of GPT-4o (Output Reasoning, Python)}%
\label{lst:Correct Case of GPT-4o (Output Reasoning, Python)}%
\begin{lstlisting}[language=Python]
def f(s: str) -> str:    
    arr = list(s.strip())
    arr.reverse()
    return ''.join(arr)

def check(candidate):
    assert candidate('   OOP   ') == 'POO'

def test_check():
    check(f)

test_check()
\end{lstlisting}
\end{listing}

\begin{listing}[h!]%
\caption{Error Case of DeepseekCoder-6.7b-base (Output Reasoning, Python)}%
\label{lst:Error Case of DeepseekCoder-6.7b-base (Output Reasoning, Python)}%
\begin{lstlisting}[language=Python]
def f(s: str) -> str:    
    arr = list(s.strip())
    arr.reverse()
    return ''.join(arr)

def check(candidate):
    assert candidate('   OOP   ') == '   POO'

def test_check():
    check(f)

test_check()
\end{lstlisting}
\end{listing}

\begin{listing}[h!]%
\caption{Correct Case of DeepseekCoder-6.7b-base (Input Reasoning, C++)}%
\label{lst:Correct Case of DeepseekCoder-6.7b-base (Input Reasoning, C++)}%
\begin{lstlisting}[language=C++]
#include<assert.h>
#include<bits/stdc++.h>
std::string f(std::string text, long size) {
    long counter = text.length();
    for (long i = 0; i < size - (size % 2); ++i) {
        text = " " + text + " ";
        counter += 2;
        if (counter >= size) {
            return text;
        }
    }
    return text;
}
int main() {
    auto candidate = f;
    assert(candidate(("7"), ("7")) == ("     7     "));
}
\end{lstlisting}
\end{listing}

\begin{listing}[h!]%
\caption{Error Case of Phi-1.5 (Input Reasoning, C++)}%
\label{lst:Error Case of Phi-1_5 (Input Reasoning, C++)}%
\begin{lstlisting}[language=C++]
#include<assert.h>
#include<bits/stdc++.h>
std::string f(std::string text, long size) {
    long counter = text.length();
    for (long i = 0; i < size - (size % 2); ++i) {
        text = " " + text + " ";
        counter += 2;
        if (counter >= size) {
            return text;
        }
    }
    return text;
}
int main() {
    auto candidate = f;
    assert(candidate(("7"), (11)) == ("     7     "));
}
\end{lstlisting}
\end{listing}
\end{document}